\documentclass{article}

 \usepackage[preprint,nonatbib]{neurips_2025}

\usepackage[utf8]{inputenc} % allow utf-8 input
\usepackage[T1]{fontenc}    % use 8-bit T1 fonts
\usepackage{hyperref}       % hyperlinks
\usepackage{url}            % simple URL typesetting
\usepackage{booktabs}       % professional-quality tables
\usepackage{amsfonts}       % blackboard math symbols
\usepackage{nicefrac}       % compact symbols for 1/2, etc.
\usepackage{microtype}      % microtypography
\usepackage[table,dvipsnames]{xcolor}         % colors
\usepackage{multirow,multicol}
\usepackage{amsmath}
\usepackage{amssymb, amsthm}
\usepackage{tabularx}
\usepackage{caption}
\usepackage{subcaption}
\usepackage[observedlatentcolor]{cv-dags}
\usepackage{wrapfig}
\usepackage{bm}
\usepackage{bbm}
\usepackage{mathtools}
\usepackage{enumitem}

\newcommand{\eg}{e.g., }
\newcommand{\ie}{i.e., }

\def\eqref#1{equation~\ref{#1}}
\def\1{\bm{1}}

\def\gC{{\mathcal{C}}}
\def\gD{{\mathcal{D}}}
\def\gE{{\mathcal{E}}}
\def\gF{{\mathcal{F}}}
\def\gG{{\mathcal{G}}}

\def\gU{{\mathcal{U}}}
\def\gV{{\mathcal{V}}}
\def\gW{{\mathcal{W}}}

\def\gY{{\mathcal{Y}}}

\def\sD{{\mathbb{D}}}
\newcommand{\E}{\mathbb{E}}

\newcommand{\R}{\mathbb{R}}

\newcommand{\aceMV}{ACE_M(V\!:\! v\! \rightarrow\! \tilde{v})}
\newcommand{\hataceMV}{\widehat{ACE}_M(V\!:\!v\! \rightarrow\! \tilde{v})}

\usepackage[backend=biber, defernumbers=true, style=nature, maxnames=99, maxcitenames=2, date=year, isbn=false, url=false, doi=true, natbib=true]{biblatex}
\title{Causality-Driven Audits of Model Robustness}

\author{%
  Nathan Drenkow$^{1,2}$
  ~~~~~William Paul$^{1,2}$~~~~~Chris Ribaudo$^1$ \\
  $^1$The Johns Hopkins University Applied Physics Laboratory \\
  Laurel, MD USA \\
  \and 
  \textbf{Mathias Unberath}$^2$ \\
  $^2$The Johns Hopkins University \\
  Baltimore, MD USA
}

\begin{document}
\maketitle

%%%%%%%%% ABSTRACT
\begin{abstract}
   Robustness audits of deep neural networks (DNN) provide a means to uncover model sensitivities to the challenging real-world imaging conditions that significantly degrade DNN performance in-the-wild.  Such conditions are often the result of multiple interacting factors inherent to the environment, sensor, or processing pipeline and may lead to complex image distortions that are not easily categorized.  When robustness audits are limited to a set of isolated imaging \textbf{effects} or distortions, the results cannot be (easily) transferred to real-world conditions where image corruptions may be more complex or nuanced. To address this challenge, we present a new alternative robustness auditing method that uses causal inference to measure DNN sensitivities to the factors of the imaging process that \textbf{cause} complex distortions. Our approach uses causal models to explicitly encode assumptions about the domain-relevant factors and their interactions.  Then, through extensive experiments on natural and rendered images across multiple vision tasks, we show that our approach reliably estimates causal effects of each factor on DNN performance using only observational domain data.  These causal effects directly tie DNN sensitivities to observable properties of the imaging pipeline in the domain of interest towards reducing the risk of unexpected DNN failures when deployed in that domain. 
\end{abstract}

%%%%%%%%% BODY TEXT
\section{Introduction}
\label{sec:intro}
A persistent challenge in the development and use of vision-based AI systems is dealing with the diversity of possible real-world operating conditions.  Safety- and cost-critical applications require robust and reliable algorithms capable of maintaining their performance across diverse conditions seen during deployment. Recent studies~\cite{Taori2020-ky, Djolonga2020-oi, Fang2023-xp, Ibrahim2022-lk, Shankar2019-nd} have shown that despite significant developments in deep learning methods, DNNs remain susceptible to performance degradation due to challenging natural imaging conditions. Robustness audits thus play a critical role in identifying the sensitivities of DNNs to these conditions before models are deployed in high-stakes applications. A compounding challenge is that the most demanding imaging conditions may arise as a result of multiple interacting factors related to the external environment (\eg, lighting, weather, shadows),  sensor (\eg, exposure, ISO, F-stop), and processing pipeline (\eg auto white balance, tone mapping, denoising). 

\begin{figure}
    \centering
    \includegraphics[width=0.8\linewidth]{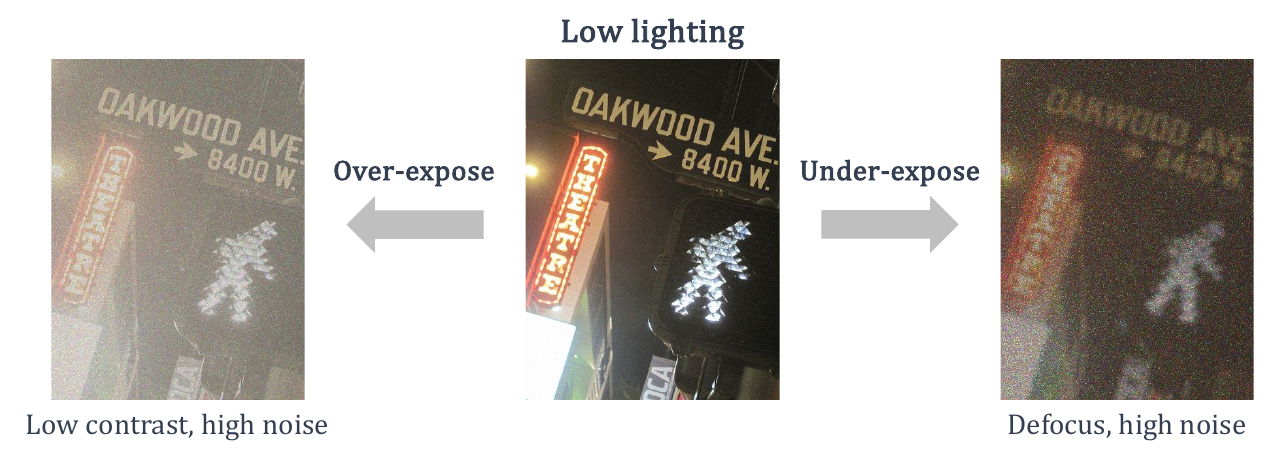}
    \caption{Illustrative example of the ``exposure triangle''.  Over- and under-exposure of the image may produce compounded effects with low contrast, low depth of field, and/or image noise.
    }
    \label{fig:exposure-triangle}
    \vspace{-0.5cm}
\end{figure}

As an illustrative example, consider the well-understood ``exposure triangle'' in natural photography where exposure time, aperture size (or F-stop), and ISO settings are adjusted to achieve the desired photographic appearance depending on lighting conditions.  In limited illumination settings, captured images may contain combinations of effects such as limited depth of field (large aperture), blurring due to sensor/object motion (long exposure times), and higher noise (high ISO to account for low photon counts). Because it is not always possible to find an ideal compromise, all three distortions may occur simultaneously and in varying degrees of severity depending on the settings  (Fig.~\ref{fig:exposure-triangle}).  Furthermore, attempting to offset one effect (\eg, exposure time) during imaging by adjusting one of the other settings (\eg, ISO) may only result in a change in the nature of the distortion (\eg, reduced motion blur but higher noise). Similarly, changes in environmental factors, such as time of day, weather, or location, will subsequently cause exposure, ISO, and F-stop settings to have varying impact on the overall image quality. Given this potential range of complexity in real-world conditions, we ask: \textbf{how can we identify the set of imaging factors that have the strongest effect on DNN robustness?}

In order to effectively audit the robustness of DNNs to identify these factors, we need the following criteria.  First, we need methods that can be grounded in domain knowledge and aligned with the types of imaging conditions expected in the target domain. This grounding and alignment ensures that results obtained from an audit will be predictive of future performance in the targeted deployment conditions.  Second, we need audits to produce interpretable and actionable insights. In this context, audits should identify sensitivity to factors that developers can observe/control in training and deployment in order to improve model robustness.  
% For instance, if a model is sensitive to changes in image exposure, developers can re-train the DNN on more diverse exposures and/or ensure the exposure setting is properly chosen given the lighting conditions during deployment. 
Finally, ideally it should be possible to conduct audits on complex real-world data whenever possible.  

The current conventional form of robustness auditing uses the common corruptions framework~\cite{Hendrycks2019-ye, Laugros2021-fd, dong2023benchmarking, He2023-jg, Mao2023-fb, Altindis2021-oo, Michaelis2019-gi} (Fig.~\ref{fig:common-corruptions-dag}) which identifies and evaluates a limited set of isolated \emph{effects} of the image generating process (IGP) (\eg, blur, noise, contrast). While this framework is capable of evaluating a range of conditions, it does not provide a sound theoretical framework for analyzing DNN sensitivities to complex conditions that occur when multiple imaging factors are compounded in real-world data (such as in the illustrative example above).  In this way, many of these evaluations are not grounded by specific domain knowledge and lead to findings that are difficult to translate to real-world settings.  Similarly, by focusing on effects of the IGP, it is challenging to derive actionable insights for understanding or predicting how DNN performance may change as a result of changes to settings of \emph{causes} of the imaging conditions. Lastly, by evaluating isolated imaging corruptions, these audits must always rely on either simulated or additional manual annotation of conditions to produce the requisite data for evaluation.

To address these limitations, we propose an alternative first-of-its-kind robustness auditing framework that analyzes DNN sensitivities with respect to the factors in the IGP that \emph{cause} image distortions. 
Our audits are aligned with the imaging domain by using causal models to explicitly encode prior knowledge and assumptions about imaging factors and their interactions affecting image quality (such as the ``exposure triangle'' example shown in Fig.~\ref{fig:cdra-dag}). We can draw upon extensive knowledge of imaging pipelines (\eg, natural~\cite{Karaimer2016-oi, Delbracio2021-ym, Szeliski2022-ck} or medical CT~\cite{Withers2021-pb, Kilim2022-ac} domains) to identify relevant domain-specific imaging factors and compose them to form causal graphs for targeted domains of interest (Sec.~\ref{sec:methods}).  Using the tools of causal inference, our causality-driven robustness audits (CDRA) provide a theoretical framework for identifying DNN sensitivities to individual \emph{causes} of the IGP using real-world domain data containing a range of natural imaging conditions resulting from multiple causal factors. 
% Our method generates interpretable and actionable insights tied to observable factors of the imaging process that enable improved data collection/augmentation, DNN design/training, and DNN generalization performance prediction.

\noindent\textbf{Contributions:~}
We make the following contributions:
\begin{itemize}[leftmargin=*]
\itemsep0em
    \item We introduce our novel CDRA framework that enables analysis of model sensitivities to complex imaging domains comprised of multiple interacting factors
    \item We show that task DNN robustness to individual imaging factors depends heavily on properties of the domain encoded via causal models
    \item We demonstrate empirically that CDRA can be applied effectively to a wide range of possible domains and complex imaging conditions using observational data sampled directly from these domains
    \item We show that CDRA is itself robust to errors in the specification of the causal model
\end{itemize}

\begin{figure*}[t!]
    \centering
    \begin{subfigure}[b]{0.45\linewidth}
        \centering
        \resizebox{.75\linewidth}{!}{%
        \input{dags/common-corruptions-dag}
        }%
        \caption{Analysis of \textit{effects}: Common corruptions~\cite{Hendrycks2019-ye}}
        \label{fig:common-corruptions-dag}
    \end{subfigure}
    \begin{subfigure}[b]{0.45\linewidth}
        \centering
        \resizebox{0.85\linewidth}{!}{%
        \input{dags/cdra-dag}
        }%
        \caption{Analysis of \textit{causes}: \textbf{CDRA (ours)}}
        \label{fig:cdra-dag}
    \end{subfigure}
    \caption{Causal graphs representing contrasting robustness auditing paradigms: (a) Common corruptions~\cite{Hendrycks2019-ye} and (b) \textbf{CDRA (ours)}. Each arrow represents a causal relationship and DNN performance is measured given labels ($Y$), predictions ($\hat{Y}$), and the desired metric ($M = f_M(Y, \hat{Y})$).
    The common corruptions approach focuses on a subset of possible \textit{effects} (e.g., \textbf{B}lur, \textbf{C}ontrast, and \textbf{N}oise) but does not consider any interactions or co-occurrence between them (i.e., no arrows between \{\textbf{B, C, N}\}).
    In contrast, our CDRA approach enables analysis of DNN sensitivities to \textit{causes} of image distortion (i.e., \textbf{L}ighting, \textbf{E}xposure, \textbf{F}-stop, \textbf{ISO}) while using images containing complex corruptions that result from multiple interacting factors that occur in real-world domains.
    }
    \label{fig:main}
    \vspace{-0.25cm}
\end{figure*}

\section{Related work}
\label{sec:related_work}
Robustness research in deep learning for computer vision largely falls into two categories: adversarial and natural/non-adversarial robustness.  
We focus here on \textit{natural robustness} where there is no explicit attacker and performance of the DNN is measured against challenging, naturally-occurring conditions~\cite{Drenkow2021-pw}.  Robustness audits in this context are often framed around distribution shift~\cite{Taori2020-ky, Djolonga2020-oi, Lu_undated-ha, Baek2022-ze, Sun2022-yr, Ibrahim2022-lk}, out-of-distribution (OOD)~\cite{Zhao2021-gl, Wenzel2022-ah}, or common corruption robustness~\cite{Hendrycks2019-ye, Mao2023-fb, He2023-jg, Laugros2021-fd, dong2023benchmarking, Michaelis2019-gi}.

\noindent\textbf{Conventional robustness audits:~}
A majority of natural robustness auditing methods center around the common corruptions framework~\cite{Hendrycks2019-ye} which initially proposed 15 classes of corruptions each simulated \emph{independently} at five discrete levels of severity.  This approach was later expanded to other domains~\cite{He2023-jg, Dong2023-yy} and tasks~\cite{Michaelis2019-gi, Mao2023-fb, Altindis2021-oo}.  Related methods proposed a similar evaluation on a subset of common image perturbations~\cite{Laugros2021-fd, Laugros2019-lv}. The use of metrics like mean Corruption Error~\cite{Hendrycks2019-ye} or robustness score~\cite{Laugros2019-lv} are useful when distortions occur independently but are not suitable for measuring the impact of individual factors of the IGP in the presence of other interacting factors. The discrete and independent nature of these simulated corruptions and robustness evaluations prevents generalizing the results to domains where multiple corruptions often co-occur, thus increasing the risk of unexpected DNN failures in deployment.

\noindent\textbf{Causal inference for computer vision:~}
Causal inference methods have shown promise for analyzing~\cite{Jones2023-wv, Castro2020-ic, Pavlak2023-ea, Butcher2021-ns, Vlontzos2022-rc} and generating~\cite{Drenkow2023-mt, Garcea2021-he} complex, large-scale datasets. We adopt a similar perspective here by focusing on specifying and using causal models of the IGP for \textit{robustness evaluation} in contrast to methods that use causal models for representation learning~\cite{Scholkopf2021-xy, Lopez-Paz2016-gr, Zhang2022-ma, Qin2021-up, Veitch2021-vx, Xu2023-wr, Chalupka2014-jc, Zhang2021-kh, Ilse2020-gu}.

\section{Methods}
\label{sec:methods}
We now provide details on auditing task DNN robustness through the lens of causal inference. 
% We first describe the specification of causal models for domain-specific image generating processes. 
% We then provide a method for analyzing DNN sensitivities to factors of the image generating process using the tools of causal inference. 

\subsection{Causal models of the image generating process}
\label{sub:GCM}
The basis for our robustness audit is the specification of a Graphical Causal Model (GCM) which explicitly codifies knowledge and/or assumptions about the image generating process of a domain $\gD$. Formally, we start by specifying the set of primary imaging factors $\gV$ in the domain (\eg $\{L, E, F\} \subset \gV$ in Fig.~\ref{fig:cdra-dag}). Each factor $V \in \gV$ corresponds to an observable property of the environment (\eg, time of day, weather, or location), sensor (\eg, exposure, f-stop, ISO), or other aspect of the imaging pipeline (\eg, white balance, compression) that impacts the formation and quality of the image $X$ (\eg Fig.~\ref{fig:cdra-dag}). 

The GCM is specified in the form of a directed acyclic graph (DAG) $\gG_\gD = (\gV, \gE)$ with variables $\gV$ and directed edges $\gE$ where edges encode pairwise causal relationships between these variables.  
The existence of directed edges $(U, V) \in \gE$ between any pair of factors $U, V \in \gV$ is tied explicitly to the assumptions and knowledge of the domain (\eg, relationship between how exposure time influences the ISO setting in natural photography). DAGs represent explicit and grounded hypotheses about the nature of the imaging process.  For many imaging domains, we may be able to fully specify the DAG, and in others, we can use the DAG to test targeted hypotheses about the nature of the domain.  

Under the Markov assumption, a variable in the GCM is independent of its ancestors conditioned on its direct parents ($V \perp \anc(V)~|~\pa(V)$ where $\anc(V)$ and $\pa(V)$ are the ancestors and parents of $V$ respectively).  With this assumption, any factor is determined by $V = f_V(\pa(V), U_V)$ where $f_V(\cdot)$ is the causal mechanism and $U_V$ is a latent term representing measurement noise or other exogenous stochasticity. For our causal audits, we require no explicit assumptions on the nature of $f_V$ or $U_V$.

\subsection{Causality-driven robustness auditing}
\label{sub:ace}
The objective of CDRA is to estimate the causal effects that individual factors ($V \in \gV$) of the IGP have on the performance of the DNN using image data collected under realistic conditions.
We first choose a performance metric (\eg correctness/accuracy) $M: \gY \times \gY \rightarrow \R$ for labels $\gY$ .  The prediction $\hat{Y} = f_{\hat{Y}}(X, U_X; \theta)$ is determined by the task DNN $f_{\hat{Y}}$ parameterized by $\theta$ and trained to approximate $P(Y|X)$. We assume that $\theta$ is fixed so that $f_{\hat{Y}}$ is deterministic and the latent noise term can be omitted.  We augment our causal graph by adding edges from $X \rightarrow \hat{Y}$, $X \rightarrow Y$, $\hat{Y} \rightarrow M$ and $Y \rightarrow M$ (as in Fig.~\ref{fig:cdra-dag}).  The metric calculation is given as $M = f_M(Y,\hat{Y},U_M)$, and similarly, $U_M$ can be dropped since $M$ is typically deterministic. 

Given the final DAG (including $X$ and $M$), our goal is to estimate for each factor $V$ the average causal effect ($ACE$) of that factor on $M$.  We formulate this as 
\begin{equation}
    \aceMV = \E[M|do(V\! =\! \tilde{v})] - \E[M|do(V\! =\! v)]    
\label{eq:ace}
\end{equation}
which measures the expected difference in $M$ when we intervene (indicated by $do(\cdot)$) to set $V$ to two different values $\tilde{v},v$. The notion of intervention here is akin to removing the edges in the causal graph into $V$, setting the value of $V$ to $v$, and resampling all remaining variable values according to the modified graph. This operation represents the \emph{hypothetical} case where every image had been captured with $V=v$.  

To estimate $\aceMV$, we start from the causal estimand (\eg $\E[M|do(V=v)]$) and use the structure of the causal graph to determine whether we can convert this causal estimand to a statistical/observational one.  This is the problem of \textit{identifiability} in causal inference, and if we can reduce Eq.~\ref{eq:ace} to a purely statistical estimand, then  $V$ is \textit{identifiable} and can be estimated directly from observational data. While our approach is not limited with respect to the number of factors in the causal DAG, the DAG topology and potential presence of unobserved factors can impact identifiability.  Because the sensor and/or imaging process for many applications and domains is well-understood, we can often make the assumption that all critical imaging factors affecting image quality are observable and accounted for in the GCM. Various techniques are available for identification and, in particular, backdoor/frontdoor adjustments~\cite{Pearl2009-af} are used based on the structure of the causal graphs in our experiments.  These techniques identify the particular subset of variables $\gW$ that must be adjusted for in order to prevent confounding in the $ACE$ estimates (see Appendix~\ref{app:example-adjustment} for an example of a GCM DAG with adjustment sets $\gW$ given for each variable).  
 
Once we obtain a statistical estimand, many techniques are available for estimating $\aceMV$ including traditional methods such as S-/T-/X-learners~\cite{Kunzel2019-ai}. In our experiments, we use S-learners that are designed as models $\mu(w, v) = \E\left[M|W=w, V=v\right]$ where $V$ is the targeted factor and $W$ are the adjustment variables used to prevent biased $ACE$ estimates.  Then, we estimate $\aceMV$ using the estimator $\hat{\mu}$:
\begin{equation}
    \widehat{ACE}_M(V\! : v\! \rightarrow\! \tilde{v}) = \frac{1}{\mid\gD\mid} \sum_{w \in \gD} \hat{\mu}(w, \tilde{v}) - \hat{\mu}(w, v)
\end{equation}
\noindent where $\hat{\mu}$ can be implemented with a variety of machine learning approaches. In general, the choice of technique for estimating $ACE$ will depend on the nature of the variables and their relationships in the causal DAG.

\subsection{Conducting CDRA}
\label{sub:conducting-cdra}
For CDRA, we estimate $\aceMV$ for each $V \in \gV$. When $M$ is the typical correctness metric (i.e., $\1[\hat{y} = y]$), we can interpret $\aceMV$ as the change in accuracy when imaging factor $V$ changes from $v$ to  $\tilde{v}$. In this case, we say that a DNN is robust to factor $V$ when $\aceMV$ is close to $0$.  Let $v_0$ be a nominal value for factor $V$ such that any deviation away from nominal (i.e., $|\tilde{v} - v_0| > 0$) is likely to degrade the image quality, then we say a DNN is less robust for $ACE_M(V\!:v_0\! \rightarrow\! \tilde{v}) < 0$ because the task DNN's performance is expected to decrease as a result of a change in $V$ that degrades the imaging conditions.

It is important to note that because of the identifiability process (described above) and the observability of most key imaging factors, we can often obtain an estimate $\aceMV$ purely from observational data without requiring the ability to directly modify or intervene on the actual image generation process.  This means that we can collect a real-world evaluation dataset $\gD$ under the expected imaging conditions of the domain, while still estimating isolated causal effects of each imaging factor on DNN performance. In this way, we can evaluate DNNs on complex imaging conditions and still obtain estimates of the sensitivity of performance (via $\aceMV$) to changes in observable imaging factors that \emph{cause} those conditions.

\section{Simulating complex image domains}
\label{sec:synthetic-data}
In order to show experimentally that CDRA can expose DNN sensitivities in complex imaging domains, we require evaluation data with several properties. First, we need datasets where we can precisely control the imaging factors, and their interactions, affecting image quality. Second, we need the ability to test our auditing approach across a diverse range of domains/conditions. Third, we need datasets that also still reflect the complexity of objects and scenes present in the real world.  We see that previous datasets and benchmarks~\cite{Hendrycks2019-ye, Michaelis2019-gi, Laugros2021-fd, Schmalfuss2025-us} do allow for precisely controlling factors that lead to image degradation, but do not consider interactions amongst factors.  Other approaches~\cite{Drenkow2023-mt, Ibrahim2022-lk} have introduced more complexity in simulating imaging domains, but have relied on scenes that lack aspects of real-world complexity.  Lastly, while a large number of public image benchmark datasets exist that cover a wide range of real-world scenes and conditions~\cite{Deng2009-yc, Irvin2019-jk, Chambon2024-vz, Schuhmann2022-ea}, challenging conditions are often excluded and crucial metadata about the underlying imaging factors (typically encoded in EXIF tags) is typically removed (\eg due to privacy concerns, memory constraints, collection protocol).

To address these limitations, we generate a new set of synthetic datasets with complex imaging conditions that meet the necessary criteria above.  In particular, we adapt the image corruptions from the ImageNet-C benchmark~\cite{Hendrycks2019-ye} and the causal model-based rendering from~\cite{Drenkow2023-mt} to simulate a range of hypothetical domains whereby (1) we have full ground truth of the imaging process and underlying factors, (2) we can produce imaging conditions that go beyond the complexity of existing benchmarks and approach more natural, real-world settings, and (3) we can still audit the robustness of DNNs on real-world scenes.  Using this data generation framework, we aim to assess the efficacy of our CDRA approach to uncover DNN sensitivities beyond what average performance metrics capture.

\noindent\textbf{GCMs for domain-specific image generation:~}
\label{sub:gen}
A primary benefit of the causal perspective on the IGP is that it enables a sparse factorization of complex distributions over factors that affect image quality.  When we ``invert'' this property, we gain the ability to precisely control the generation of complex image distributions which we can use for evaluating the efficacy of CDRA. These complex distributions allow for simulating a wide range of imaging conditions whereby corruptions compound to yield complex distortions.  

To generate such distributions, we first define the domain $\gD$ using the DAG $\gG_\gD = (\gV, \gE)$.  
Underlying each domain is a joint distribution over all factors and images, $P_\gG(\gV, X) = \prod_{V \in \gV} P(V | pa(V))$, which we can sample to generate evaluation datasets for our simulated domain.
In the case of generating evaluation data for these hypothetical domains, we may also need to specify a set of causal mechanisms $\gF = \{f_V(pa(V), U_V)| \forall V \in \gV \}$ and latent noise distributions that determine each factor's value during sampling. 
We use the GCM definition in conjunction with one of two corruption generation processes (\textit{compositing} and \textit{rendering}) to generate our evaluation datasets.

\begin{table}[t!]
    \centering
    \caption{Set of available corruptions as nodes in simulated GCMs.}
    \resizebox{0.5\columnwidth}{!}{%
    \begin{tabular}{c|c||c|c}
    \toprule
    Variable & Name & Variable & Name\\
    \midrule
    G & Gaussian noise & D & Defocus blur \\ 
    N & Shot noise & C & Contrast  \\
    I & Impulse noise & B & Brightness \\
    S & Speckle noise & S & Saturate \\
    G & Gaussian blur & P & Pixelate \\
    \bottomrule
\end{tabular}    
    }%
    \label{tab:corruptions}
    \vspace{-0.25cm}
\end{table}

\noindent{\textbf{Compositing:~}} 
\label{subsub:composite}
In the compositing setting (adapted from~\cite{Drenkow2023-mt}), we specify a set of image corruption functions~\cite{Hendrycks2019-ye} to be applied in sequence according to the specified GCM DAG. In this setting, variables $V$ in the GCM correspond to normalized severities for an associated set of corruption functions $\gC = \{c_V~|~V \in \gV \}$ where $c_V(x, V) = \tilde{x}$. We first sample the severity values using the causal model (i.e., $V = f_V(pa(V), U_V)$ for all $V \in \gV$).  We then apply the corresponding corruption functions to the original image by following a topological ordering from $\gG_\gD$ and using the image output by the previous function in the ordering as input to the next corruption. 

% This approach builds on the well-established common corruption framework (i.e., \cite{Hendrycks2019-ye}) but now allows for compounded corruptions that yield more complex imaging effects.  This approach is conceptually similar to the AugMix~\cite{Hendrycks2019-cx} augmentation strategy but where the corruption ordering and severity here is determined and sampled according to the causal model representing the knowledge about the domain. 
% % Visual examples of varying generated imaging conditions can be found in Figure~\ref{fig:natural-example-1} in Appendix~\ref{app:images}.

\noindent{\textbf{Rendering:~}} 
\label{subsub:render}
In this work, we also extend beyond~\cite{Drenkow2023-mt} by introducing a \textit{rendering} approach which uses the causal model to directly guide physics-based rendering.  Here, factors of the causal model correspond directly to settings of the rendering engine (i.e., Blender~\cite{Blender}).  These settings typically mirror the types of measurements that are available when capturing image datasets in real-world settings.  For instance, the GCM may contain factors such as exposure, ISO, and focal length, all of which map directly to settings in Blender.
% and similarly available in EXIF tags in real world scenarios.

To render corrupted images, we sample the factor values $\bm{v} \sim P_\gG(\gV)$ and update the Blender settings directly with these values for a single scene. Then the physics-based \verb|Cycles| engine renders the scene under the sampled conditions. The rendering process directly captures the full set of effects of the sampled factor values on the lighting, materials, scene geometry and dynamic properties of the scene.

\section{Experiments}
\label{sec:experiments}
In the following experiments, we seek to show that CDRA allows us to identify sensitivities of DNN performance to changes in imaging factors that produce complex imaging conditions. 
% As described in the previous section, we utilize simulated hypothetical domains to overcome limitations of existing public benchmarks.  
We first examine the accuracy of CDRA in domains where we can simulate complex imaging conditions \emph{and} perform interventions to compute ground truth $\aceMV$.  
We then examine the effects of misspecification of the causal graph on estimating $\aceMV$.
Lastly, we apply CDRA to additional vision tasks and imaging domains to show that it generalizes and provides useful insights beyond image classification.

\begin{figure*}[t!]
    \centering
    \begin{minipage}{\textwidth}
        \begin{minipage}{\linewidth}
            \centering
            \captionof{table}{True $ACE_{acc}(V\!:\! 0\!\rightarrow\! 1)$ ($\%$) per variable in a subset of GCMs in our compositing-generated datasets (left side). Single factor GCMs (right side) correspond to the common corruptions framework (Fig.~\ref{fig:common-corruptions-dag}) which did not evaluate multiple interacting factors. Values $<0$ indicate that DNN accuracy decreases due to an increase in corruption severity of $V$ in the GCM. $ACE_{acc}$ close to $0$ (or $>0$) is an indicator of greater robustness. See App.~\ref{app:full-ace} for true and estimated $ACE_{acc}$ across all 10 GCMs \& factors.}
            \resizebox{\linewidth}{!}{%
            \begin{tabular}{c|ccccc|ccccc|ccccc}
\toprule
GCM & \multicolumn{5}{c}{0} & \multicolumn{5}{c}{1} & \multicolumn{5}{c}{2} \\
DNN / Factor & G & IN & N & P & S & C & G & GN & IN & P & B & D & G & GN & N \\
% DNN &  &  &  &  &  &  &  &  &  &  &  &  &  &  &  &  &  &  &  &  \\
\midrule
ConvNext-B & -5.3 & -4.5 & -3.4 & -6.9 & -8.1 & -22.5 & -2.4 & -4.9 & -12.5 & -2.4 & -0.64 & -8.3 & -3.3 & -1.8 & -6.3 \\
ResNet50 & -7.0 & -7.0 & -3.6 & -5.3 & -11.5 & -27.2 & -4.8 & -5.9 & -12.4 & -3.9 & 0.04 & -12.4 & -4.7 & -4.1 & -13.6 \\
Swin-B & -4.8 & -4.9 & -4.0 & -9.1 & -7.3 & -17.1 & -3.8 & -4.7 & -9.9 & -8.3 & 0.44 & -9.1 & -3.2 & -1.8 & -4.6 \\
\bottomrule
\end{tabular}

% \begin{tabular}{c|ccccc|ccccc}
% \toprule
% GCM & \multicolumn{5}{c}{0} & \multicolumn{5}{c}{1} \\
% DNN / Factor & G & IN & N & P & S & C & G & GN & IN & P \\
% \midrule
% ConvNext-B & -5.3 & -4.5 & -3.4 & -6.9 & -8.1 & -22.5 & -2.4 & -4.9 & -12.5 & -2.4 \\
% ResNet50 & -7.0 & -7.0 & -3.6 & -5.3 & -11.5 & -27.2 & -4.8 & -5.9 & -12.4 & -3.9 \\
% Swin-B & -4.8 & -4.9 & -4.0 & -9.1 & -7.3 & -17.1 & -3.8 & -4.7 & -9.9 & -8.3 \\
% \bottomrule
% \end{tabular}
            \begin{tabular}{||c|c|c|c|c}
\toprule
\multicolumn{5}{||c}{Single Factor GCMs}\\
 G & GN & IN & N & P \\
\midrule
 -8.2 & -9.0 & -12.4 & -10.0 & -9.5 \\
 -10.6 & -16.7 & -29.1 & -19.0 & -12.8 \\
 -9.3 & -8.8 & -10.8 & -9.8 & -9.4 \\
\bottomrule
\end{tabular}
            }%
            \label{tab:true-ace}        
        \end{minipage} \\
        \begin{minipage}[ht]{0.62\textwidth}
            \centering
            \begin{subfigure}[b]{0.33\linewidth}
                \centering
                \includegraphics[width=\linewidth]{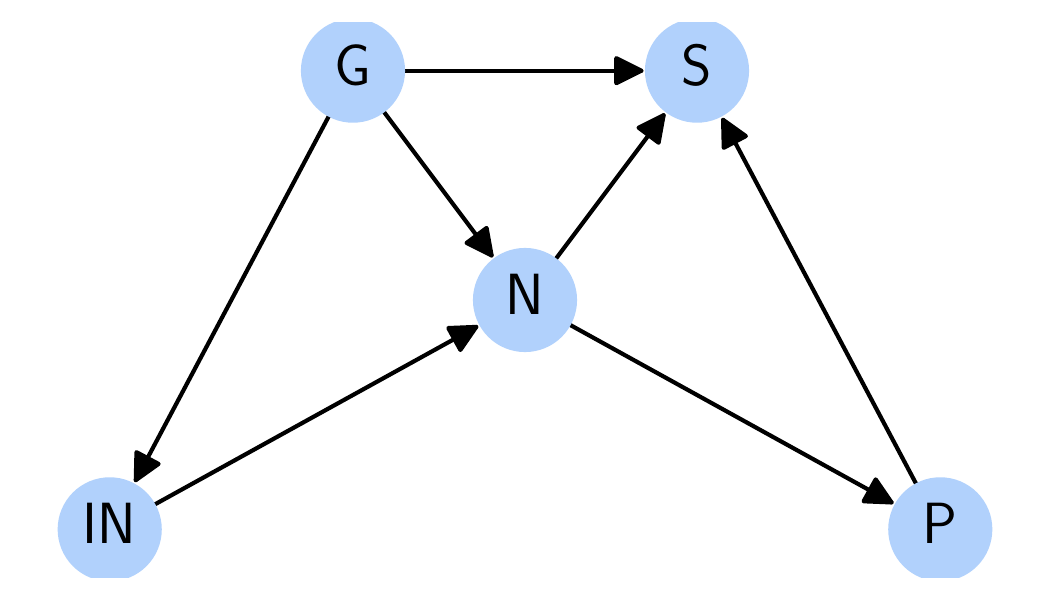}
                \caption{GCM 0}
            \end{subfigure} 
            \hfill
            \begin{subfigure}[b]{0.33\linewidth}
            \centering
                \includegraphics[width=\linewidth]{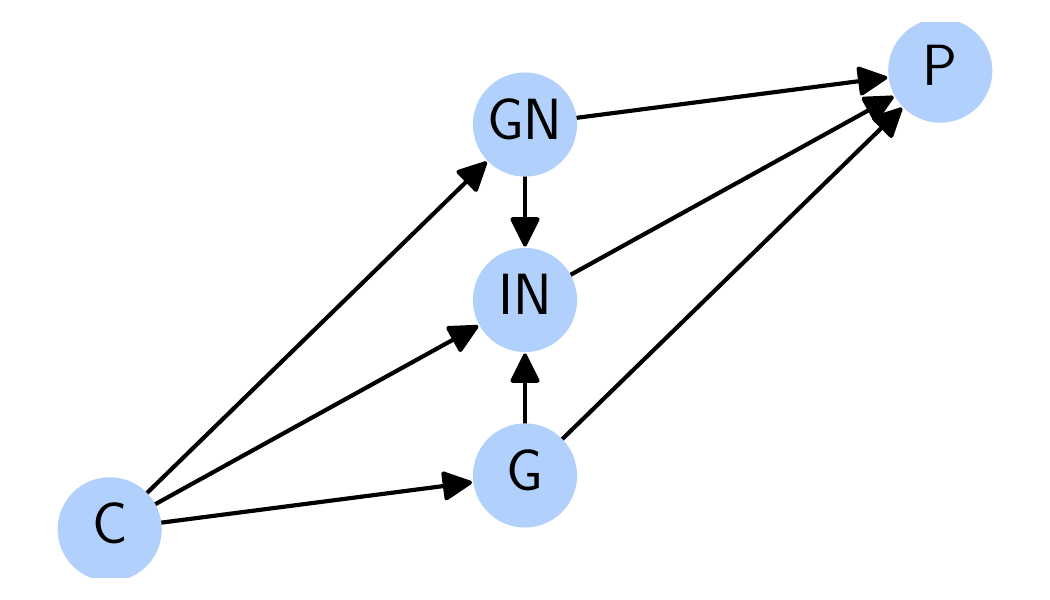} \caption{GCM 1}
            \end{subfigure} 
            \hfill
            \begin{subfigure}[b]{0.31\linewidth}
            \centering
                \includegraphics[width=\linewidth]{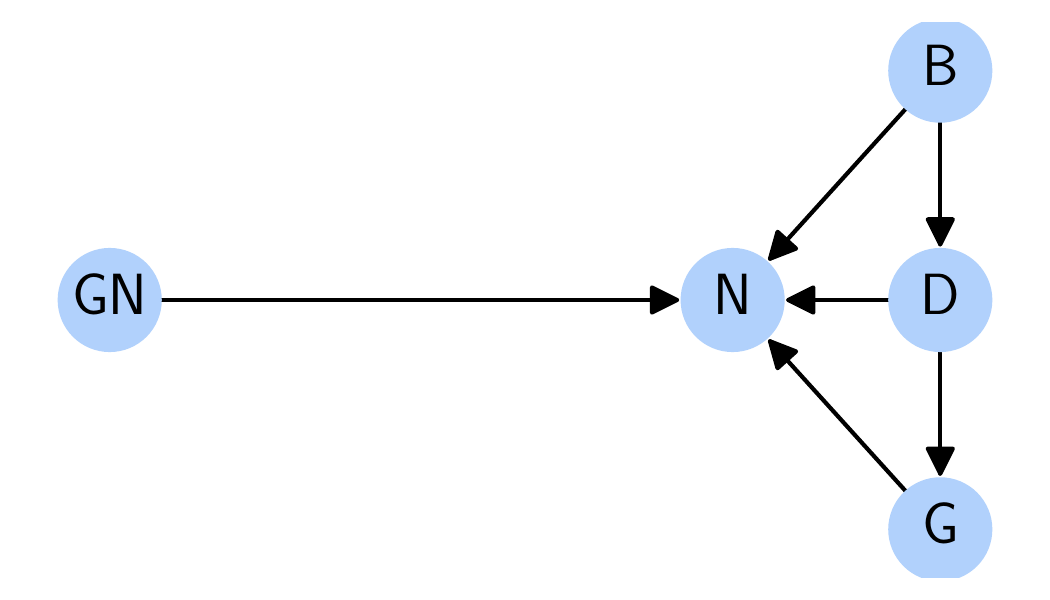} \caption{GCM 2}
            \end{subfigure} 
            \captionof{figure}{\textbf{DAGs for random GCMs (see also Table~\ref{tab:true-ace})}  Images are rendered according to the corruption severities sampled from the distribution underlying each model.}
            \label{fig:random-dags}
        \end{minipage}
        \hfill
        \begin{minipage}{0.35\linewidth}
            \centering
            \captionof{table}{Comparison of $\Delta_{ACE}$ ($\%$) averaged over all GCM factors. (Full table in Appendix~\ref{app:dag-errors})}
            \resizebox{\linewidth}{!}{%
                \begin{tabular}{c|ccc}
% & \multicolumn{3}{c}{$\Delta_{ACE}$} \\
\toprule
% & \multicolumn{3}{c}{DNN} \\
GCM  & ConvNext-B & ResNet50 & Swin-B \\
\midrule
0 & 1.0 & 0.83 & 0.70 \\
1 & 1.3 & 0.84 & 0.87 \\
2 & 0.84 & 1.1 & 0.85 \\
\midrule
\multicolumn{1}{r|}{Mean (all GCMs)} & 0.91 & 0.76 & 0.79\\
\multicolumn{1}{r|}{Std (all GCMs)} & 0.79 & 0.74 & 0.72 \\
\bottomrule
\end{tabular}
            }%
            \label{tab:error-ace-by-gcm}
            \vspace{-0.25cm}
        \end{minipage}
    \end{minipage}
    \vspace{-0.5cm}
\end{figure*}

\subsection{Auditing DNNs on diverse image domains}
\label{sub:exp-cdra}
In order to demonstrate the generalizability of CDRA, we start by running audits over a set of complex imaging domains. For each simulated hypothetical domain, we evaluate a set of task models on all images in the domain.  We then apply CDRA to compute $\aceMV$ for each factor/cause $V$ in the domain GCM. 

\textbf{GCM sampling and image generation:~}
We first specify a set of possible corruption functions $\gC$ that will act as proxies for observable variables $V \in \gV$ in the domain simulated via the causal model.  
Given $\gC$ (see Table~\ref{tab:corruptions}), we generate a set of GCMs by sampling a DAG of $N$ variables $\gV$ (with each $V \in \gV$ associated with a corruption function $c_V \in \gC$) and with probability $p((V_i,V_j) \in \gE) = 0.5$ of a directed edge from $V_i \rightarrow V_j$ being formed between any pair of variables in the DAG.  Also associated with each variable in the GCM is a set of conditional probability distributions (CPDs) that encode the relationship ($P(V\mid\pa(V))$) between the sampled value of the variable ($V$) and its parents ($\pa(V)$). For our experiments, we sample 10 GCMs with $N=5$ factors to allow for our imaging conditions to be sufficiently complex and in order to test whether CDRA can accurately estimate $ACE$ values under such challenging conditions.
We generate data for each simulated domain (example DAGs shown in Figure~\ref{fig:random-dags}, full set in Appendix~\ref{app:full-ace}). 

% Given the GCM variables, corruption functions, and DAG, we simulate sampling from the domain as follows.  
For each image in the ImageNet-Val dataset~\cite{Deng2009-yc}, we sample the severity values for each $V \in \gV$ from the distribution underlying the GCM and then apply the corresponding corruptions following the compositing procedure outlined in Section~\ref{sec:synthetic-data}. We limit severity values to $\{0, 1, 2\}$ (with $0$ corresponding to $c_V(x, 0) = x$ and severity $2$ included to allow for more variability in imaging conditions) in order to ensure that images corrupted with multiple functions will still be interpretable. For each simulated domain, we generate 50k samples with varying levels of corruption consistent with the GCM (see Appendix~\ref{app:sample-images} for sample visualizations).

\noindent{\textbf{Ground truth $ACE$:~}}
To obtain ground truth $\aceMV$ for comparison against the $\hataceMV$ estimates, we need to evaluate task DNNs on image data consistent with the interventional distributions in Eq.\ref{eq:ace} (\ie $P(M|do(V=v)), P(M|do(V=\tilde{v}))$).
% We generate additional variants of each GCM's dataset where we directly intervene on targeted variables.  
For each GCM, we intervene by removing edges into $V$ and setting $V$ to the corresponding intervention value (\ie $do(V=v)$). We then resample the other variables according to the GCM's distribution and render a new dataset with those sampled values. The ground truth $\aceMV$ is computed as the difference in task DNN accuracy on the pair of interventional dataset variants.
For our experiments, we calculate ground truth $ACE_M(V\!: 0\! \rightarrow\! 1)$ with $M$ as accuracy which measures the true change in task DNN accuracy when the corruption severity associated with $V$ goes from $0$ (no corruption) to $1$ (presence of corruption). For comparison, we also compute ground truth $ACE$ in the common corruptions framework using a set of single node GCMs where corruptions are applied individually and independently.  

\noindent{\textbf{Task DNNs:~}}
For all experiments, we evaluate a set of task DNNs that cover a range of architectural design patterns, sizes, and levels of performance.  In particular, we evaluate ResNet50~\cite{he2016deep}, ConvNext-B~\cite{liu2022convnet}, and Swin-B Transformer~\cite{Liu2021-pj} on each simulated GCM domain.  All models are pretrained on ImageNet~\cite{Deng2009-yc} and were not exposed to corrupted images during training, ensuring that DNN performance is not biased towards any GCM and corresponding dataset.

\noindent{\textbf{CDRA setup:~}}
We conduct CDRA for each GCM and task model pair.  We (initially) assume full knowledge of the GCM DAG for estimating causal effects.  We use an S-Learner based on Random Forest regression for estimating $\aceMV$ which is capable of capturing non-linear interactions in the data while limiting model bias in the $ACE$ estimates. For each variable $V$ and given $M\!\coloneqq\!\1[\hat{Y} = Y]$ (\ie correctness), we estimate the $\hataceMV$ and also compute the ground truth $\aceMV$ as described above.  We compute the ACE error as 
$\Delta_{ACE} = \left|\hataceMV - \aceMV)\right|$.  

% \begin{table}[t!]
%     \centering
%     \caption{Mean Top-1 accuracy ($\%$) for DNNs on datasets associated with GCMs in Fig.~\ref{fig:random-dags}.  See Appendix~\ref{app:full-ace} for all GCM results.}
%     \resizebox{.7\linewidth}{!}{%
%         \input{tables/mean-acc}
%     }%
%     \label{tab:mean-acc}
%     \vspace{-0.25cm}
% \end{table}

\noindent{\textbf{Results:~}}
In Table~\ref{tab:true-ace}, we see the per-node ground truth $ACE_{acc}$ results computed for a subset of the simulated domains associated with the corresponding GCM DAGs in Figure~\ref{fig:random-dags}.  The full set of these $ACE$ results across all GCMs can be found in Appendix~\ref{app:full-ace}.  
For comparison, the right side of Table~\ref{tab:true-ace} shows the $ACE_{acc}$ under the common corruptions framework that assumes corruptions occur in isolation and without interaction.
First, we observe the general trend across GCMs and task DNNs that increasing the corruption severity of any node by one has a moderate negative effect on task DNN performance (average $ACE=-5\%$) and in some specific cases that impact is severe ($ACE=-22\%$). 
For a given imaging factor in each GCM (including the single factor domains), we observe $ACE$ values vary across DNNs showing that each DNN is sensitive to different factors depending on the domain properties given by the GCM.  
Similarly, we see that the magnitude of the $ACE$ values for a given imaging factor may vary significantly from one domain GCM to another (\eg see $G$ in GCMs 0, 1 and the single-factor case).  
We also observe similar average accuracies (Table~\ref{tab:app-mean-acc-full}) that reflect the general lack of robustness in the task DNNs, despite the fact that the average corruption severity per GCM factor is $<1$ across the dataset.  

To ensure that the $ACE$ estimates are accurate, we compute $\Delta_{ACE}$ for each GCM individually using the ground truth $ACE$ obtained as described in Section~\ref{sub:exp-cdra}. The results are summarized in Table~\ref{tab:error-ace-by-gcm} and show that across all GCMs, the average error in $ACE$ is less than $1\%$.
% indicating that accurate $ACE$ estimates can be obtained directly from domain data.

\noindent{\textbf{Discussion:~}}
These results show that under complex imaging conditions, CDRA allows us to effectively estimate how individual factors of the image generating process influence the task DNNs performance. These findings are consistent across a range of simulated domains with diverse and complex GCMs.  Whereas mean accuracy (the conventional robustness metric) is only able to summarize model performance on the dataset, CDRA provides deeper insights into how properties of the domain that affect image quality more directly influence DNN behavior.  The fact that there is variance in $ACE$ values occurs across DNNs, imaging factors, and GCMs illustrates how observed DNN robustness depends heavily on the nature of the imaging domain and the types of conditions it produces.  The low $\Delta_{ACE}$ values also show that we can accurately recover $ACE_M$ from evaluations using observational data.

A primary and critical implication of these results is that CDRA enables analyzing fine-grained aspects of DNN behavior directly on domain imaging data collected under complex and diverse conditions.  While previous work has largely focused on computing mean performance metrics over images affected by isolated corruptions, CDRA allows for evaluation on image data containing compounded corruptions that are reflective of real-world imaging conditions. Differences in $ACE_{acc}$ values between the factors in the sampled GCMs (Table~\ref{tab:true-ace}) and the single-factor cases (Table~\ref{tab:true-ace}) illustrate that evaluation results obtained via the common corruptions framework may not be predictive of DNN behavior in more complex imaging domains. This underscores the benefit of CDRA as it enables isolating sensitivities to imaging factors via $ACE_{acc}$ while relaxing the data requirements to allow for direct evaluation on complex domain data.

\subsection{Assessing sensitivity of CDRA to GCM DAG misspecification}
\label{sub:exp-dag-err}

The previous experiment demonstrated that CDRA could provide accurate, deep, and fine-grained insights into DNN performance.  However, the results were contingent on the assumption that full knowledge of the GCM DAG is available. For many domains, this assumption is reasonable given that knowledge of the imaging process can be translated into an accurate specification of the DAG.  
However, other imaging domains may be complex, and misspecification of the causal DAG may result in errors in $ACE$ estimates. In many domains, it may be possible to identify at minimum the key variables as well as a limited set of known interactions (such as exposure, aperture, and ISO in the ``exposure triangle'' example earlier), but the presence/absence of other edges in the DAG may be ambiguous.  

In this experiment, we test for edge-related DAG misspecifications as follows.  For each GCM, we run CDRA using assumed DAGs that differ from the true domain DAG.  To create DAG errors, we randomly sample up to $N_E \in \{1, 2, 4\}$ edges to add or delete from the DAG (where the number of edges added may be limited by the constraint that the graph must remain a DAG).  We then run CDRA to estimate $\aceMV$ for each $V$ in the GCM using the misspecified DAG. We re-run this process 5 times with different selections of the missing/added edges. As in Sec.~\ref{sub:exp-cdra}, we compute the error ($\Delta_{ACE}$) to assess the impact of the DAG-related errors (Table~\ref{tab:app-error-ace-edge-errors-full}).  We compute a \textit{residual error} by subtracting the baseline error for each DAG (see Table~\ref{tab:error-ace-by-gcm}) from the new estimates to show how $ACE$ estimation error is impacted by the DAG specification errors.

\begin{figure}[t!]
    \centering
    \begin{minipage}[b!]{\columnwidth}
        \centering
        \captionof{table}{\textbf{Effect of $N_E$ DAG edge errors on $ACE$ estimation.} Residual error is the deviation from the baseline estimation error (in $\%$) when the DAG is correctly specified. Errors are averaged over all GCMs and corresponding factors. Close to 0 is best.}
        \resizebox{0.6\linewidth}{!}{%
            \begin{tabular}{c|ccc|ccc|ccc}
& \multicolumn{9}{c}{Missing Edges} \\
\toprule
\textbf{DNN} & \multicolumn{3}{c}{ConvNext-B} & \multicolumn{3}{c}{ResNet50} & \multicolumn{3}{c}{Swin-B} \\
\textbf{$N_E$} & 1 & 2 & 4 & 1 & 2 & 4 & 1 & 2 & 4 \\
\midrule
% 0 & -0.10 & -0.28 & -0.21 & -0.17 & -0.27 & -0.31 & -0.06 & -0.15 & -0.15 \\
% 1 & -0.27 & 0.03 & 0.55 & -0.14 & 0.06 & 0.38 & -0.12 & 0.24 & 0.82 \\
% 2 & 0.06 & 0.11 & 0.31 & 0.15 & 0.24 & 0.64 & 0.06 & 0.12 & 0.33 \\
% 3 & 0.19 & 0.31 & 0.49 & 0.21 & 0.41 & 0.55 & 0.19 & 0.37 & 0.55 \\
% 4 & 0.08 & 0.09 & 0.06 & 0.09 & 0.06 & -0.04 & 0.09 & 0.07 & 0.00 \\
% 5 & -0.04 & 0.02 & 0.08 & -0.05 & 0.16 & 0.32 & -0.01 & 0.14 & 0.26 \\
% 6 & 0.24 & 0.50 & 0.77 & 0.15 & 0.35 & 0.80 & 0.22 & 0.53 & 0.91 \\
% 7 & 0.12 & 0.16 & 0.23 & -0.08 & -0.12 & -0.02 & 0.02 & 0.08 & 0.01 \\
% 8 & 0.29 & 0.64 & 1.6 & 0.21 & 0.44 & 0.83 & 0.29 & 0.60 & 1.3 \\
% 9 & 0.01 & 0.06 & 0.46 & 0.08 & 0.22 & 0.64 & 0.05 & 0.06 & 0.39 \\
% \midrule
Mean & 0.06 & 0.17 & 0.44 & 0.04 & 0.16 & 0.38 & 0.07 & 0.21 & 0.44 \\
Std & 0.50 & 0.79 & 1.3 & 0.45 & 0.73 & 1.0 & 0.44 & 0.78 & 1.2 \\
\bottomrule
\end{tabular} 
        } \\% 
        \resizebox{0.6\linewidth}{!}{%
            \begin{tabular}{c|ccc|ccc|ccc}
& \multicolumn{9}{c}{Additional Edges} \\
\toprule
% \textbf{DNN} & \multicolumn{3}{c}{ConvNext-B} & \multicolumn{3}{c}{ResNet50} & \multicolumn{3}{c}{Swin-B} \\
% \textbf{$N_E$} & 1 & 2 & 4 & 1 & 2 & 4 & 1 & 2 & 4 \\
% \midrule
% 0 & -0.01 & -0.02 & -0.02 & 0.01 & -0.01 & -0.00 & 0.00 & 0.00 & 0.01 \\
% 1 & 0.06 & 0.11 & 0.10 & 0.05 & 0.12 & 0.10 & 0.07 & 0.12 & 0.10 \\
% 2 & 0.01 & 0.00 & -0.07 & 0.04 & 0.06 & -0.03 & 0.02 & 0.03 & -0.03 \\
% 3 & 0.02 & 0.04 & 0.04 & 0.03 & 0.04 & -0.00 & 0.04 & 0.07 & 0.06 \\
% 4 & -0.00 & 0.01 & 0.01 & -0.02 & -0.04 & -0.02 & -0.03 & -0.05 & -0.03 \\
% 5 & -0.03 & -0.02 & -0.04 & -0.02 & -0.01 & -0.07 & -0.07 & -0.05 & -0.02 \\
% 6 & 0.00 & -0.03 & -0.03 & -0.01 & -0.04 & -0.03 & 0.01 & -0.01 & -0.01 \\
% 7 & 0.12 & 0.11 & 0.14 & 0.03 & 0.01 & 0.04 & 0.07 & 0.06 & 0.09 \\
% 8 & 0.06 & 0.07 & 0.17 & 0.04 & 0.04 & 0.17 & 0.01 & 0.01 & 0.11 \\
% 9 & 0.03 & 0.05 & 0.04 & 0.05 & 0.06 & 0.02 & 0.06 & 0.07 & 0.05 \\
% \midrule
Mean & 0.03 & 0.03 & 0.04 & 0.02 & 0.02 & 0.02 & 0.02 & 0.03 & 0.03  \\
Std & 0.11 & 0.13 & 0.13 & 0.10 & 0.13 & 0.14 & 0.11 & 0.13 & 0.11 \\
\bottomrule
\end{tabular}
        }%
        \label{tab:error-ace}
    \end{minipage}
    \vspace{-0.5cm}
\end{figure}

\noindent{\textbf{Results:~}}
The primary results for testing robustness to DAG misspecification can be found in Table~\ref{tab:error-ace} (with all detailed GCM results in Appendix~\ref{app:dag-errors}).  We find in both sets of results that CDRA is robust to misspecifications in the GCM DAG with $\Delta_{ACE}$ changing by less than $1\%$ on average from the baseline error even when the DAG differs by a substantial number of edges.  The relatively small changes in $\Delta_{ACE}$ may be attributed to a couple properties of CDRA.  
First, under the Markov assumption from Sec.~\ref{sec:methods}, the GCM DAG encodes assumed/known conditional independencies between variables such that the effects of edge errors are relatively localized due to these independence assumptions. 
Second, the identifiability process (Sec.~\ref{sec:methods}) may include/exclude incorrect imaging factors in the process of estimating $\aceMV$ when edge errors occur.  However, the estimator itself (e.g., S-learner) may be robust to minor errors in these cases and still capable of producing accurate $ACE$ estimates.

\noindent{\textbf{Discussion:~}}
We can see from this experiment that CDRA is able to provide detailed insights about DNN performance on complex image data even in the case that the assumed causal DAG includes edge-related errors.  In general, we expect that identifying the set of key variables in the DAG is reasonable, but this experiment acknowledges that accurately identifying all relationships between these variables may be difficult and shows that CDRA can tolerate a small number of errors.  

Furthermore, since CDRA can be run efficiently (relative to the cost of evaluating DNNs on the image data), we can often test several DAG hypotheses as a means of determining whether $ACE$ estimates are sensitive to edge errors when the true DAG is not fully known. In our experiment, we observed a low variance in $\Delta_{ACE}$ indicating that for multiple misspecified hypotheses for the same true DAG, the $ACE$ estimates were still relatively close to the true value.  These results were consistent across the full range of GCMs we evaluated against and indicate that CDRA generalizes well across diverse domains and potential errors.

\subsection{CDRA for additional vision tasks}
\label{sub:exp-tasks}
Lastly, we have shown in Experiments 1, 2 that we can accurately estimate $ACE$ (even in the presence of DAG specification errors), so we now use a similar experimental setup show how CDRA can generate deeper robustness insights for additional vision tasks and models.  We use data \textit{rendered} with Blender and GCMs specified for sampling the underlying physics-based rendering settings.

\subsubsection{Object-Centric Learning}
\label{subsub:ocl}
\begin{figure*}[t!]
    \centering    
    \begin{subfigure}[b]{0.38\linewidth}
        \centering
        \includegraphics[width=.85\textwidth]{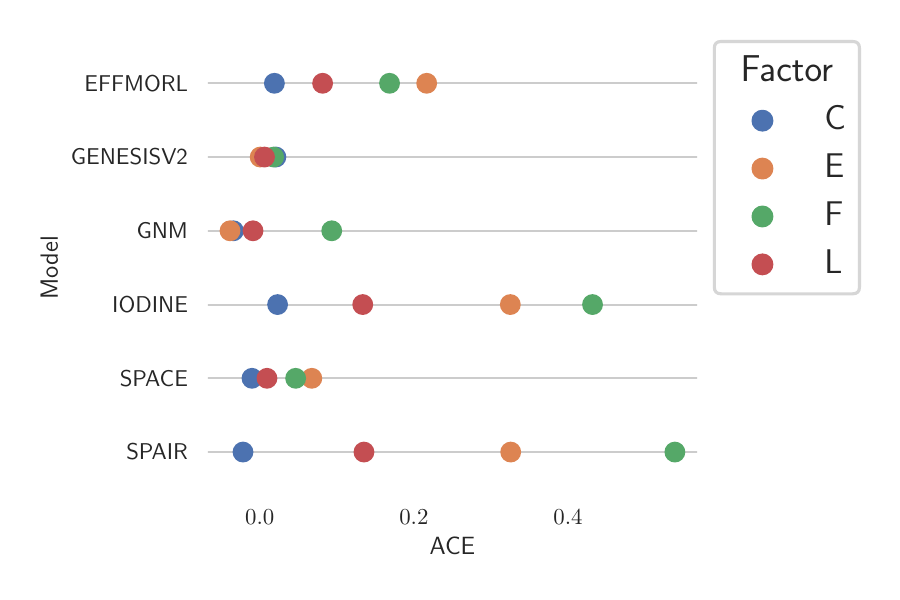}
        \caption{}
        \label{fig:ocl-ace}
        \vspace{-0.1cm}
    \end{subfigure}
    \begin{subfigure}[b]{0.38\linewidth}
        \centering
        \includegraphics[width=.8\textwidth]{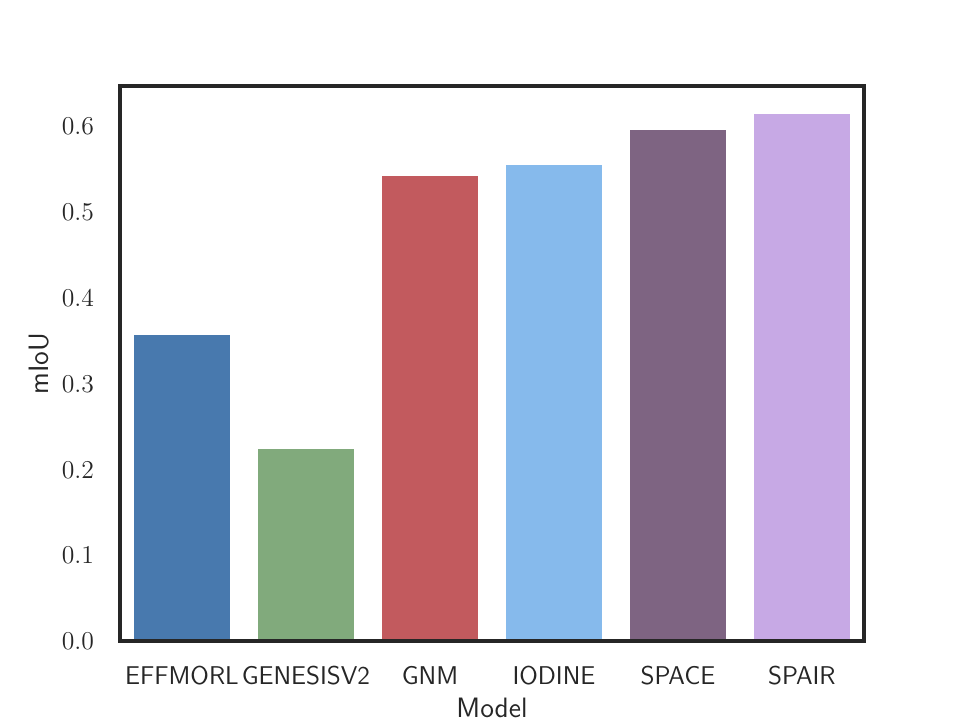}
        \caption{}
        \label{fig:ocl-miou}
        \vspace{-0.1cm}
    \end{subfigure}
    \begin{subfigure}[b]{0.2\linewidth}
        \centering
        \resizebox{.9\linewidth}{!}{%
        \input{dags/ocl-of-dag}
        }%
        \caption{}
        \label{fig:ocl-flow-scm}
        \vspace{-0.1cm}
    \end{subfigure}
    \caption{(a) $ACE_{mIoU}$ by factor in the underlying DAG and (b) Average $mIoU$ by model for corrupted images generated according to the GCM in Figure~\ref{fig:ocl-flow-scm}. Larger $ACE_{mIoU}$ indicates lower robustness of the OCL model to changes in the corresponding factor. (c) GCM DAG used to generate the CLEVR and MOVi-C eval datasets. \{\textbf{L}: Lighting, \textbf{E}: Exposure, \textbf{F}: F-stop, \textbf{C}: Render cycles\}}
    \vspace{-0.25cm}
\end{figure*}

We first consider applying CDRA for an emerging class of vision algorithms for doing object-centric learning (OCL).  
We render corrupted versions of the CLEVR~\cite{Johnson2017-po} dataset consisting of photo-realistic synthetically-rendered images of simple objects with varying colors, sizes, and materials.  
Given the causal model in Fig.~\ref{fig:ocl-flow-scm}, we use the rendering method of Sec.~\ref{sec:synthetic-data} to generate evaluation data under a wider range of imaging conditions than those found in the original CLEVR benchmark (see App.~\ref{app:sampling_corruptions} for more details).

We evaluate several baseline OCL algorithms including EffMORL~\cite{Emami2021-cw}, GENESISv2~\cite{Engelcke2021-se}, GNM~\cite{Jiang2020-ps}, IODINE~\cite{Greff2019-wm}, SPACE~\cite{Lin2020-hp}, and SPAIR~\cite{Crawford2019-wl}. All models were trained on ``clean'' images similar to the original CLEVR benchmark.  We measure task performance using mean Intersection over Union ($mIoU$) between the predicted segmentation masks and ground truth.  We apply CDRA and estimate the $ACE_{mIoU}$ for each factor of the GCM on the task performance of the model.

\noindent{\textbf{Results:}}
Figure~\ref{fig:ocl-ace} shows that causal effects expose sensitivities that are not obvious when looking at average performance alone as in Figure~\ref{fig:ocl-miou}. In particular, SPAIR and SPACE are the top performing models and have similar $mIoU$, yet SPACE appears to be more robust to changes in IGP factors as evidenced by the lower $ACE$ values for factors of the GCM. 

\noindent{\textbf{Discussion:}}
These results emphasize the value of CDRA since the audit is performed directly on the synthetic domain data rather than trying to predict domain performance based on evaluation of pre-defined IID distortions.  These results further underscore that both mean performance and $ACE$ estimates are collectively necessary since $mIoU$ measures the performance of the DNN directly while $ACE$ estimates sensitivity of the metric to changes in factor values. This is particularly evident in the case where GENESISv2 shows low $ACE$ values indicating high robustness to all factors of the IGP, yet the $mIoU$ is also low indicating that the model performs poorly on average across all conditions. Similarly, SPACE and SPAIR achieve high $mIoU$, yet SPACE appears to be more robust due to lower magnitude $ACE$ values across most factors.

\subsubsection{Optical Flow}
\label{subsub:flow}
We also apply CDRA to assess the robustness of optical flow (OF) methods for natural distortions.  As in the OCL case, we find that even when top-performing baselines achieve similar average endpoint error (EPE), CDRA exposes sensitivities to IGP factors that differ between models. More detail on the experimental design and results are found in Appendix~\ref{app:optical-flow}.

\section{Discussion and Conclusion}
\label{sec:discussion}
We present here a novel perspective on robustness auditing which uses causal inference to measure the sensitivity of DNN performance to \emph{causes} of distortion in the image generating process. 
We find that even when average performance is similar between DNN models, the $ACE$ estimates may differ measurably depending on the domain GCM and thus CDRA yields more granular insights into how model performance may change as a function of individual causes of image distortion. 

Our approach is not free of limitations.  First, our method is based on the ability to specify a plausible GCM DAG.  While we believe domain knowledge is often sufficient to specify reasonable models, we recognize that verifying the accuracy of those models may be challenging.
We also acknowledge that our experiments use simulated domains modeling only some real-world dependencies but did not address the question of how to verify SCMs for \textit{specific} real-world domains.  While CDRA can be applied to real-world data without modification to the method itself, we were compelled to rely on simulated data due to the lack of public benchmarks containing sufficient metadata from the image capture process (such as camera exposure, aperture, F-stop, ISO settings, or other EXIF tag information). Our focus in this work was on evaluating the efficacy of our CDRA approach (where simulated ground truth data was necessary), and we believe that methods for specifying and aligning SCMs to specific domains of interest is a fruitful direction for future research. 

Overall, our CDRA method paves the way for enabling DNN robustness audits directly on domain-specific data containing complex, challenging imaging conditions. By measuring DNN sensitivities relative to causes in the IGP, the derived insights are actionable and can be used to develop new mitigation methods including targeted data collection/augmentation, architecture design, or GCM-driven adaptation.

% REFRENCES
\printbibliography[title=References]

\appendix
\section{Full ACE results}
\label{app:full-ace}
\begin{figure*}[t!]
    \begin{subfigure}[b]{0.3\linewidth}
        \centering
        \includegraphics[width=\linewidth]{figures/appendix/dag-00-convnext.pdf}
        \caption{GCM 0}
    \end{subfigure}
    \begin{subfigure}[b]{0.3\linewidth}
    \centering
        \includegraphics[width=\linewidth]{figures/appendix/dag-01-convnext.pdf} \caption{GCM 1}
    \end{subfigure}
    \begin{subfigure}[b]{0.3\linewidth}
    \centering
        \includegraphics[width=\linewidth]{figures/appendix/dag-02-convnext.pdf} \caption{GCM 2}
    \end{subfigure} \\
    \begin{subfigure}[b]{0.3\linewidth}
    \centering
        \includegraphics[width=\linewidth]{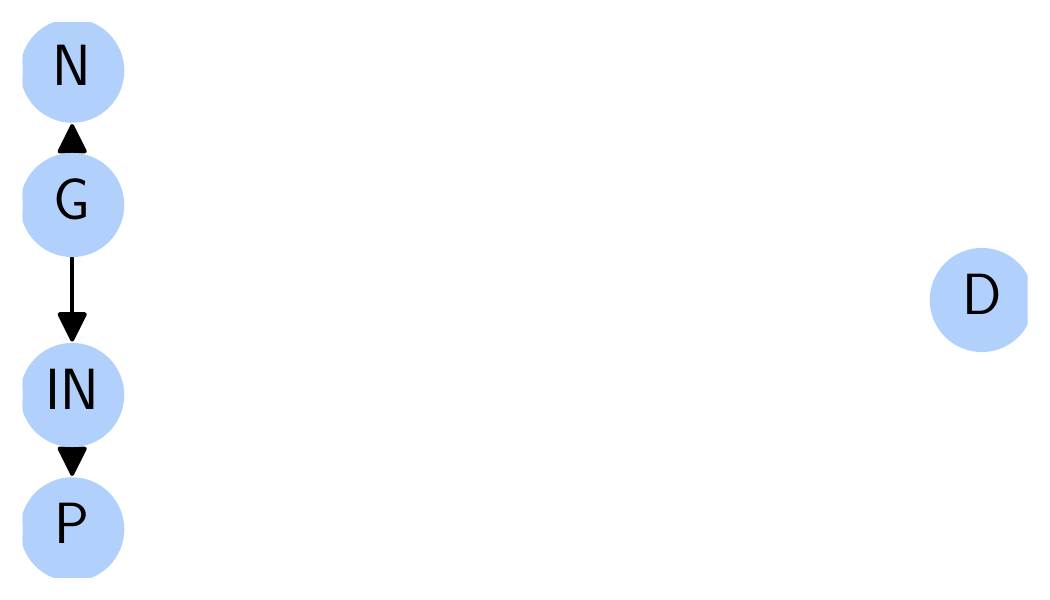}
        \caption{GCM 3}
    \end{subfigure}
    \begin{subfigure}[b]{0.3\linewidth}
    \centering
        \includegraphics[width=\linewidth]{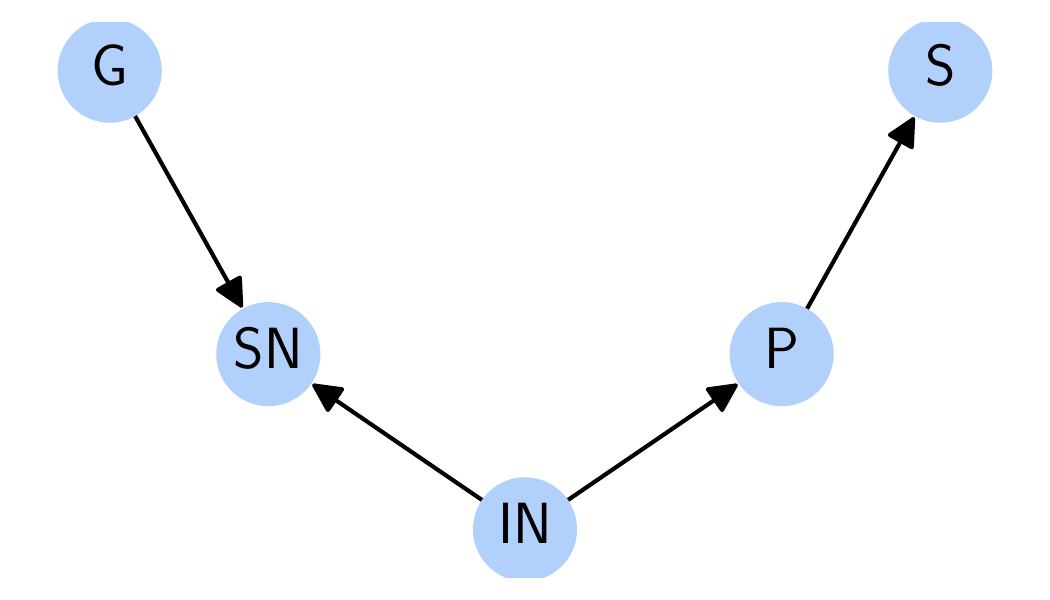} \caption{GCM 4}
    \end{subfigure}
    \begin{subfigure}[b]{0.3\linewidth}
    \centering
        \includegraphics[width=\linewidth]{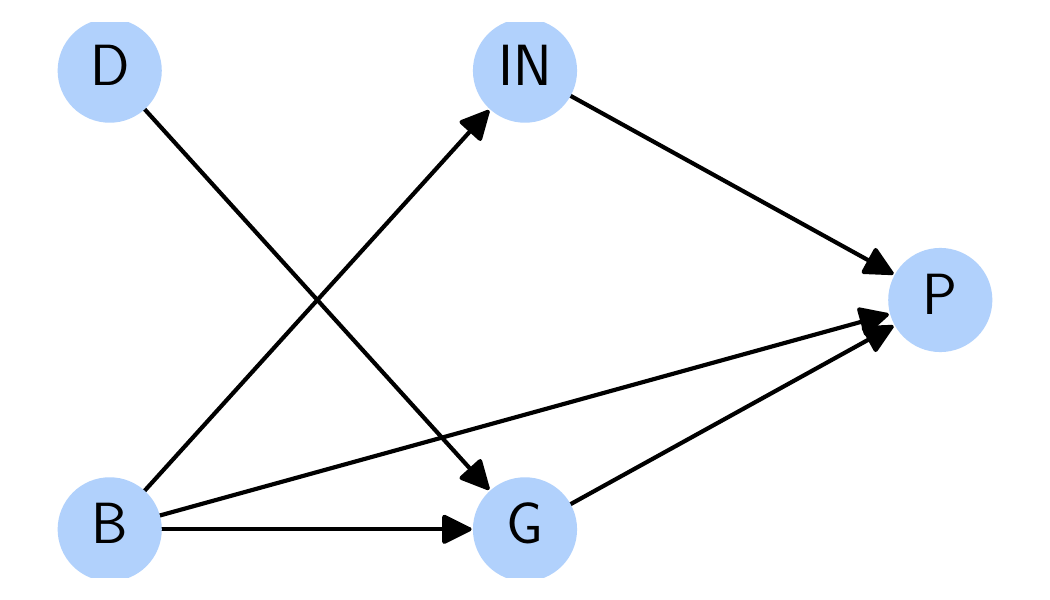} \caption{GCM 5}
    \end{subfigure} \\
    \begin{subfigure}[b]{0.3\linewidth}
    \centering
        \includegraphics[width=\linewidth]{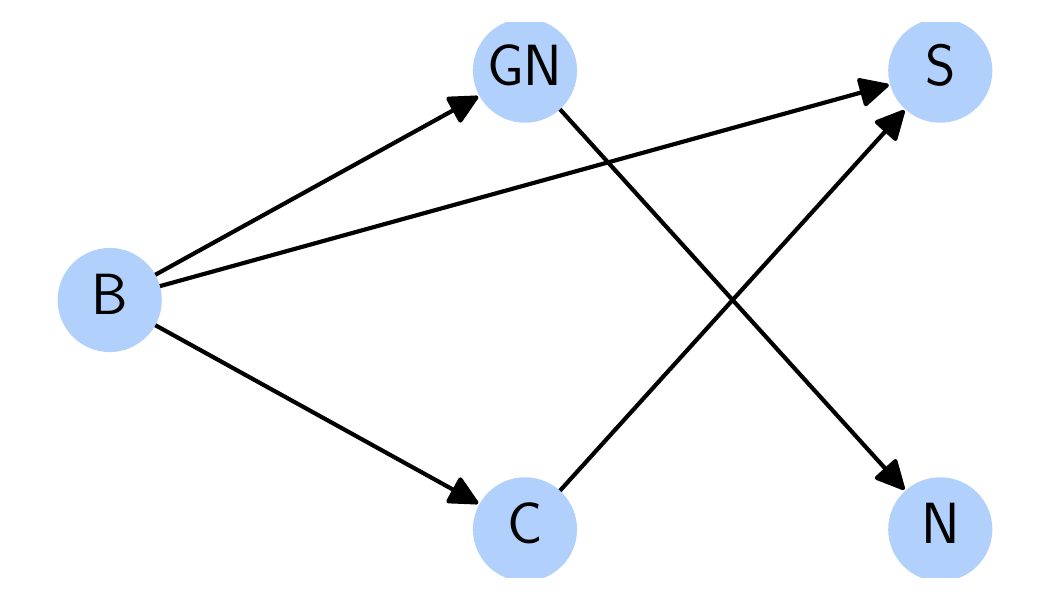}
        \caption{GCM 6}
    \end{subfigure}
    \begin{subfigure}[b]{0.3\linewidth}
    \centering
        \includegraphics[width=\linewidth]{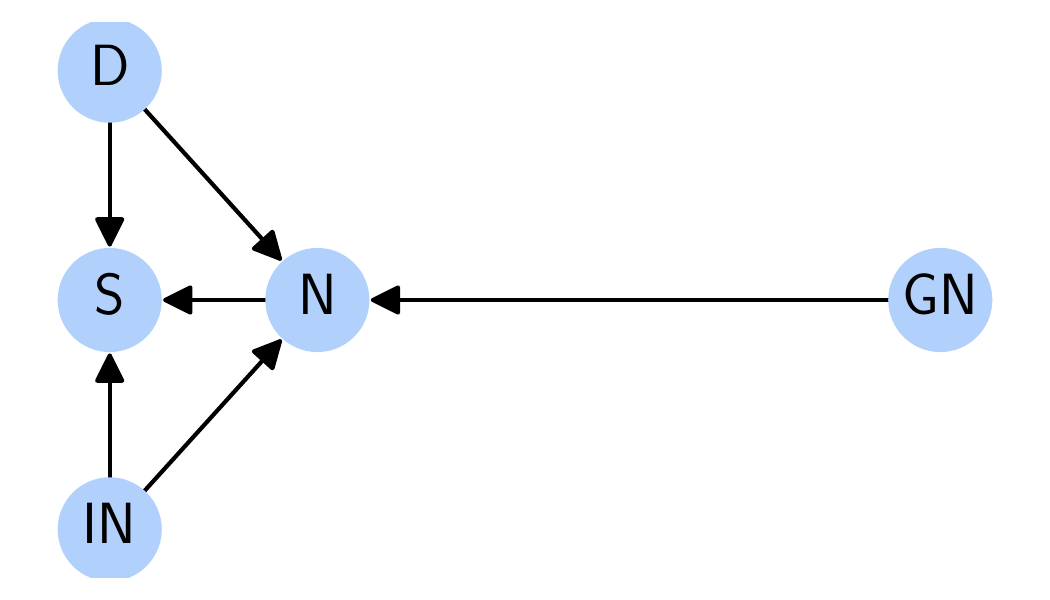} \caption{GCM 7}
    \end{subfigure}
    \begin{subfigure}[b]{0.3\linewidth}
    \centering
        \includegraphics[width=\linewidth]{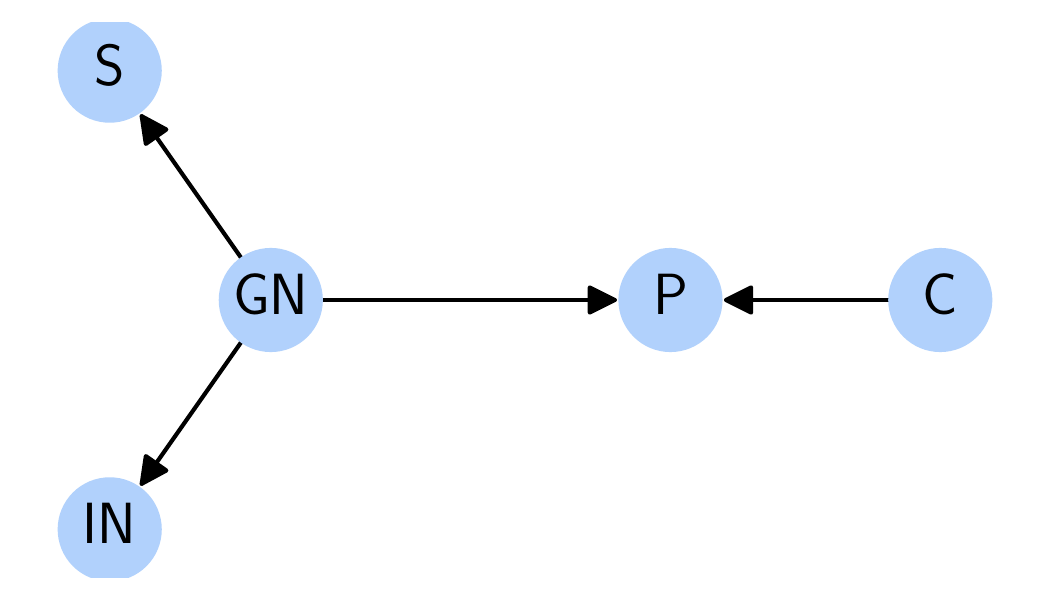} \caption{GCM 8}
    \end{subfigure} \\
    \begin{subfigure}[b]{\linewidth}
    \centering
        \includegraphics[width=0.3\linewidth]{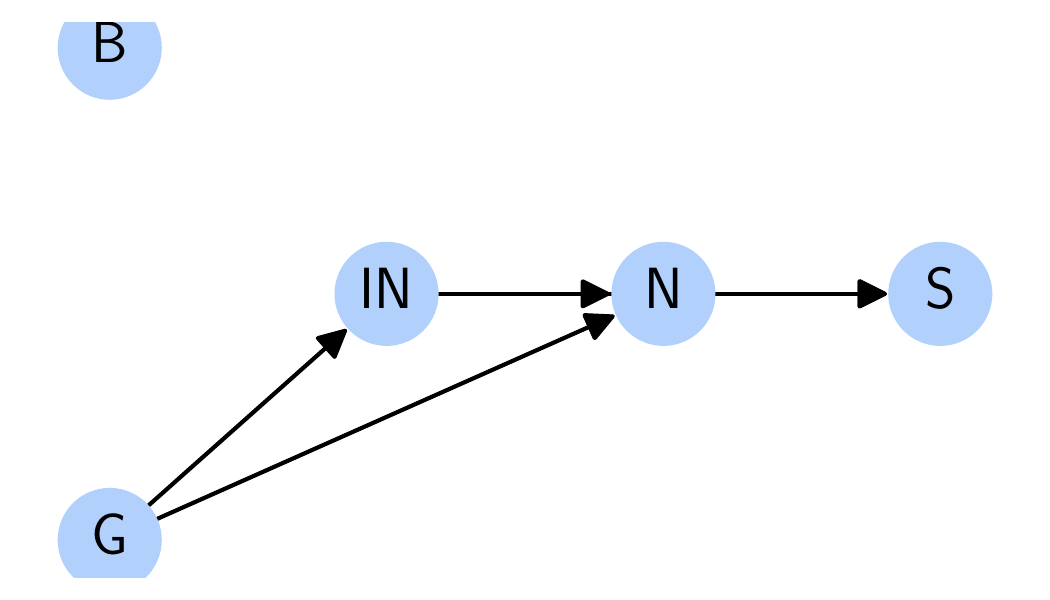} \caption{GCM 9}
    \end{subfigure}
    \caption{\textbf{DAGs for the randomly generated GCMs used to produce the data and results of Table~\ref{tab:app-est-ace-full}.} Edges from each factor in $\gV$ to $X$ and between $\{X,Y,\hat{Y},M\}$ are omitted for visual clarity.}
    \label{fig:app-random-dags-full}
\end{figure*}

As described in Section~\ref{sub:exp-cdra}, we evaluated DNNs on datasets generated using each GCM (see Figure~\ref{fig:app-random-dags-full}).  For each set of results, we run CDRA to estimate the per variable $\aceMV$.  The full set of results for all GCMs and imaging factors can be seen in Table~\ref{tab:app-est-ace-full}. Average error ($\Delta_{ACE}$) is found in Table~\ref{tab:error-ace-by-gcm}. We also show the per-GCM mean Top-1 accuracy in Table~\ref{tab:app-mean-acc-full} as well as the single node Top-1 accuracy for level 1 corruption severity in Table~\ref{tab:app-single-factor-mean-acc}.

These results provide several insights (consistent with those discussed in the main manuscript).  First, we see that mean accuracy alone gives little information about what factors affecting image quality have the strongest effect on DNN accuracy. The mean accuracies per GCM indicate that under compounded corruptions, task DNN accuracy is significantly degraded.  However, from mean accuracy alone, we cannot interpret which factors of the imaging process may be driving this substantial decrease.  In contrast, CDRA is able to accurately identify the sensitivities of DNNs to specific imaging factors grounded by knowledge or informed assumptions about the imaging domain (in the form of the causal DAG).  

Second, we see that the value of the individual $ACE$ values is strongly linked to the topology of the DAG.  For example, the presence of impulse noise ($IN$) occurs in a majority of the GCMs, and across these models, $ACE$ values range from $\approx\!-13$ up to $\approx\!-3$. We also observe that $ACE_{acc}$ for the single factor GCMs (which represent the current state-of-practice) is not predictive of $ACE_{acc}$ for the more complex GCMs (\eg compare $IN$ in Table~\ref{tab:app-single-factor-ace-full} with $IN$ in Table~\ref{tab:app-true-full-ace}).  Since the error analysis shows that our estimates $\hataceMV$ are close to the ground truth values, we can see that the single factor GCMs are in general not informative of DNN sensitivities in more complex real-world domains. These results further underscore that when knowledge of the specific imaging domain is available, it can be used to effectively estimate fine-grained DNN sensitivities.  

Lastly, we re-emphasize that we can run CDRA directly on complex image data and still obtain accurate estimates of the $\aceMV$ values.  Each GCM in this experiment produces a wide diversity of imaging conditions and compounded corruptions that prior work is able to evaluate effectively.

\begin{table*}[thbp!]
    \centering
    \caption{\textbf{True $ACE_{acc}(V\!:\! 0\! \rightarrow\! 1)$ $(\%)$ calculated for each GCM, imaging factor, and task DNN.}  Each value represents the expected change in accuracy as a result of increasing the corruption severity associated with the corresponding variable. Note how $ACE$ changes for different nodes as a function of the DAG topology. Values close to $0$ (or $>0$) indicate higher robustness.}
    \resizebox{\linewidth}{!}{%
        \begin{tabular}{c|ccccc|ccccc|ccccc}
\toprule
GCM & \multicolumn{5}{c}{0} & \multicolumn{5}{c}{1} & \multicolumn{5}{c}{2} \\
DNN / Factor & G & IN & N & P & S & C & G & GN & IN & P & B & D & G & GN & N \\
\midrule
ConvNext-B & -5.3 & -4.5 & -3.4 & -6.9 & -8.1 & -22.5 & -2.4 & -4.9 & -12.5 & -2.4 & -0.64 & -8.3 & -3.3 & -1.8 & -6.3 \\
ResNet50 & -7.0 & -7.0 & -3.6 & -5.3 & -11.5 & -27.2 & -4.8 & -5.9 & -12.4 & -3.9 & 0.04 & -12.4 & -4.7 & -4.1 & -13.6 \\
Swin-B & -4.8 & -4.9 & -4.0 & -9.1 & -7.3 & -17.1 & -3.8 & -4.7 & -9.9 & -8.3 & 0.44 & -9.1 & -3.2 & -1.8 & -4.6 \\
\bottomrule
GCM & \multicolumn{5}{c}{3} & \multicolumn{5}{c}{4} & \multicolumn{5}{c}{5} \\
DNN / Factor & D & G & IN & N & P & G & IN & P & S & SN & B & D & G & IN & P \\
\midrule
ConvNext-B & -5.3 & -3.4 & -5.0 & -2.9 & -4.2 & -4.1 & -4.6 & -12.6 & -9.6 & -5.5 & -1.4 & -5.4 & -0.35 & -3.5 & -6.4 \\
ResNet50 & -8.0 & -5.9 & -8.6 & -4.1 & -0.26 & -4.7 & -7.6 & -13.6 & -11.0 & -8.6 & -1.7 & -7.3 & 0.22 & -6.3 & -8.2 \\
Swin-B & -5.7 & -2.9 & -5.1 & -3.7 & -6.2 & -3.3 & -4.7 & -15.2 & -8.4 & -4.8 & -1.1 & -4.7 & -0.46 & -4.2 & -6.9 \\
\bottomrule
GCM & \multicolumn{5}{c}{6} & \multicolumn{5}{c}{7} & \multicolumn{5}{c}{8} \\
DNN / Factor & B & C & GN & N & S & D & GN & IN & N & S & C & GN & IN & P & S \\
\midrule
ConvNext-B & -8.5 & -28.9 & -10.8 & -19.3 & -9.4 & -13.8 & -3.5 & -6.3 & -2.9 & -9.5 & -19.2 & -12.9 & -12.8 & -5.7 & -7.8 \\
ResNet50 & -4.8 & -28.0 & -12.6 & -12.4 & -10.6 & -18.8 & -5.0 & -8.8 & -5.3 & -11.6 & -23.2 & -11.2 & -12.0 & -6.3 & -10.5 \\
Swin-B & -5.6 & -17.0 & -9.9 & -10.5 & -10.1 & -12.3 & -3.4 & -5.3 & -3.3 & -8.7 & -14.0 & -12.1 & -10.9 & -10.3 & -8.5 \\
\bottomrule
GCM & \multicolumn{5}{c}{9} & \multicolumn{10}{c}{} \\
DNN / Factor & B & G & IN & N & S & \multicolumn{10}{c}{} \\
\midrule
ConvNext-B & 0.30 & -7.5 & -7.7 & -6.7 & -7.1 & \multicolumn{10}{c}{} \\
ResNet50 & -0.15 & -9.8 & -12.5 & -9.9 & -9.3 & \multicolumn{10}{c}{} \\
Swin-B & 0.30 & -6.6 & -6.5 & -6.3 & -6.6 & \multicolumn{10}{c}{} \\
\bottomrule
\end{tabular}
    }%
    \label{tab:app-true-full-ace}
\end{table*}

\begin{table*}[t!]
    \centering
    \caption{\textbf{Estimated $\widehat{ACE}_{acc}(V\! :\!0\! \rightarrow\! 1)$ $(\%)$ calculated for each GCM and factor}  Each value represents the expected change in accuracy as a result of increasing the corruption severity associated with the corresponding variable.  Note how $ACE$ changes for different nodes as a function of the DAG topology. Values close to $0$ (or $>0$) indicate higher robustness.}
    \resizebox{\linewidth}{!}{%
        \begin{tabular}{c|ccccc|ccccc|ccccc}
\toprule
GCM & \multicolumn{5}{c}{0} & \multicolumn{5}{c}{1} & \multicolumn{5}{c}{2} \\
DNN / Factor & G & IN & N & P & S & C & G & GN & IN & P & B & D & G & GN & N \\
\midrule
ConvNext-B & -5.7 & -6.6 & -3.5 & -8.2 & -7.4 & -21.8 & -2.1 & -4.2 & -9.8 & -1.9 & -2.0 & -8.7 & -2.7 & -2.6 & -7.0 \\
ResNet50 & -7.1 & -8.9 & -4.7 & -5.6 & -12.2 & -26.2 & -3.7 & -5.1 & -12.1 & -4.1 & -1.7 & -10.9 & -3.9 & -3.2 & -14.1 \\
Swin-B & -4.8 & -6.1 & -3.8 & -10.3 & -6.7 & -16.6 & -3.4 & -3.9 & -8.8 & -7.9 & -1.4 & -9.0 & -3.8 & -2.4 & -5.1 \\
\bottomrule
GCM & \multicolumn{5}{c}{3} & \multicolumn{5}{c}{4} & \multicolumn{5}{c}{5} \\
DNN / Factor & D & G & IN & N & P & G & IN & P & S & SN & B & D & G & IN & P \\
\midrule
ConvNext-B & -6.0 & -3.0 & -5.3 & -2.0 & -5.0 & -4.5 & -5.7 & -12.8 & -8.8 & -4.2 & -0.56 & -5.5 & -0.05 & -4.0 & -7.5 \\
ResNet50 & -9.3 & -5.3 & -8.5 & -4.4 & -0.17 & -5.5 & -7.9 & -14.0 & -11.0 & -8.3 & -0.75 & -7.5 & 0.47 & -6.2 & -8.4 \\
Swin-B & -5.5 & -2.5 & -5.2 & -2.6 & -7.5 & -4.0 & -4.6 & -14.0 & -7.8 & -4.3 & -0.64 & -5.0 & -0.73 & -3.4 & -6.6 \\
\bottomrule
GCM & \multicolumn{5}{c}{6} & \multicolumn{5}{c}{7} & \multicolumn{5}{c}{8} \\
DNN / Factor & B & C & GN & N & S & D & GN & IN & N & S & C & GN & IN & P & S \\
\midrule
ConvNext-B & -6.8 & -29.7 & -11.9 & -18.8 & -7.6 & -13.9 & -2.7 & -6.8 & -4.0 & -10.3 & -17.7 & -10.0 & -12.1 & -6.7 & -8.0 \\
ResNet50 & -2.9 & -28.4 & -11.5 & -12.7 & -10.6 & -18.1 & -4.2 & -8.9 & -4.3 & -10.7 & -22.5 & -8.5 & -12.1 & -6.9 & -11.1 \\
Swin-B & -4.8 & -18.2 & -9.6 & -10.5 & -8.4 & -12.0 & -2.9 & -5.2 & -3.9 & -8.0 & -12.6 & -9.3 & -10.3 & -11.1 & -8.8 \\
\bottomrule
GCM & \multicolumn{5}{c}{9} & \multicolumn{10}{c}{} \\
DNN / Factor & B & G & IN & N & S & \multicolumn{10}{c}{} \\
\midrule
ConvNext-B & 0.20 & -6.7 & -6.6 & -5.9 & -6.9 & \multicolumn{10}{c}{} \\
ResNet50 & -0.33 & -10.7 & -11.5 & -8.9 & -10.1 & \multicolumn{10}{c}{} \\
Swin-B & -0.82 & -5.6 & -5.1 & -6.5 & -5.8 & \multicolumn{10}{c}{} \\
\bottomrule
\end{tabular}
    }%
    \label{tab:app-est-ace-full}
\end{table*}

\begin{table*}[t!]
    \centering
    \caption{\textbf{True $ACE_{acc}$ for GCMs with a single factor}.  Each column represents the $ACE(V\!:\!0\!\rightarrow\!1)$ for a DAG with a single variable and edge pointing to the image $X$ (as in the common corruptions framework of Figure~\ref{fig:common-corruptions-dag}). }
    \resizebox{0.9\textwidth}{!}{%
    \begin{tabular}{r|c|c|c|c|c|c|c|c|c|c}
\toprule
% & \multicolumn{10}{c}{Single Factor GCMs} \\
DNN / GCM & B & C & D & G & GN & IN & N & P & S & SN \\
\midrule
ConvNext-B & -4.4 & -7.2 & -14.4 & -8.2 & -9.0 & -12.4 & -10.0 & -9.5 & -5.8 & -9.4 \\
ResNet50 & -4.7 & -14.1 & -19.6 & -10.6 & -16.7 & -29.1 & -19.0 & -12.8 & -11.8 & -16.8 \\
Swin-B & -4.4 & -7.0 & -15.6 & -9.3 & -8.8 & -10.8 & -9.8 & -9.4 & -6.4 & -9.3 \\
\bottomrule
\end{tabular}
    }%
    \label{tab:app-single-factor-ace-full}
\end{table*}

\begin{table}[htbp!]
    \centering
    \caption{Mean Top-1 Accuracy for each of the GCM datasets.}
    \resizebox{0.5\linewidth}{!}{%
        \begin{tabular}{c|ccc}
\toprule
GCM / DNN & ConvNext-B & ResNet50 & Swin-B \\
\midrule
0 & 60.9 & 39.8 & 59.7 \\
1 & 53.4 & 33.4 & 55.8 \\
2 & 68.2 & 48.1 & 67.2 \\
3 & 62.4 & 42.5 & 60.9 \\
4 & 58.2 & 35.8 & 57.4 \\
5 & 66.0 & 49.4 & 64.2 \\
6 & 51.6 & 34.2 & 59.4 \\
7 & 60.2 & 35.3 & 60.9 \\
8 & 48.8 & 27.9 & 51.4 \\
9 & 64.7 & 41.2 & 64.6 \\
\bottomrule
Clean & 83.7 & 76.1 & 83.2 \\
\bottomrule
\end{tabular}
    }%
    \label{tab:app-mean-acc-full}
\end{table}

\begin{table*}[htbp!]
    \centering
    \caption{Mean Top-1 Accuracy for single factor GCMs (\ie  common corruptions framework - Fig.~\ref{fig:common-corruptions-dag}) with corruption severity 1.}
    \resizebox{0.8\linewidth}{!}{%
        \begin{tabular}{rcccccccccc}
\toprule
DNN / Factor & B & C & D & G & GN & IN & N & P & S & SN \\
\midrule
ConvNext-B & 79.3 & 76.5 & 69.3 & 75.5 & 74.7 & 71.3 & 73.7 & 74.2 & 77.9 & 74.3 \\
ResNet50 & 71.4 & 62.1 & 56.5 & 65.5 & 59.4 & 47.0 & 57.1 & 63.3 & 64.3 & 59.3 \\
Swin-B & 78.7 & 76.1 & 67.5 & 73.9 & 74.4 & 72.3 & 73.4 & 73.8 & 76.8 & 73.9 \\
\bottomrule
\end{tabular}
    }%
    \label{tab:app-single-factor-mean-acc}
\end{table*}

\section{Full DAG misspecification results}
\label{app:dag-errors}
Table~\ref{tab:app-error-ace-edge-errors-full} contains the full set of results illustrating how $ACE$ estimation error changes as a function of misspecifications of the GCM DAG.  Both tables show how much the estimation error deviates from the baseline estimation error and each cell is an average over all factors in the GCM.  The results show that misspecifications of the DAG contribute less than $1\%$ additional error to the $ACE$ estimates.  The mean and standard deviation of this residual error increase slightly as the degree of DAG misspecification error increases, but the total $ACE$ error in the most extreme cases is still relatively small.

\begin{table}[thbp!]
    \centering
    \caption{Baseline $\Delta_{ACE}$ ($\%$) calculated for each GCM and averaged across all imaging factors in the GCM DAG.}
    \resizebox{0.5\linewidth}{!}{%
    \begin{tabular}{c|ccc}
% & \multicolumn{3}{c}{$\Delta_{ACE}$} \\
\toprule
% & \multicolumn{3}{c}{DNN} \\
GCM  & ConvNext-B & ResNet50 & Swin-B \\
\midrule
0 & 1.0 & 0.83 & 0.70 \\
1 & 1.3 & 0.84 & 0.87 \\
2 & 0.84 & 1.1 & 0.85 \\
3 & 0.71 & 0.62 & 0.74 \\
4 & 0.89 & 0.54 & 0.73 \\
5 & 0.57 & 0.36 & 0.45 \\
6 & 1.2 & 0.81 & 0.97 \\
7 & 0.70 & 0.68 & 0.51 \\
8 & 1.2 & 0.97 & 1.2 \\
9 & 0.61 & 0.85 & 0.89 \\
\midrule
\multicolumn{1}{r|}{Mean (all GCMs)} & 0.91 & 0.76 & 0.79\\
\multicolumn{1}{r|}{Std (all GCMs)} & 0.79 & 0.74 & 0.72 \\
\bottomrule
\end{tabular}
    }%
    \label{tab:app-error-ace-by-gcm-full}
\end{table}

\begin{figure*}[t!]
    \captionof{table}{\textbf{Effect of DAG errors on $ACE$ estimation for $N_E$ edge errors.} Calculated as the deviation from the baseline estimation error (in $\%$) when the DAG is correctly specified. Close to 0 is best. (See Table~\ref{tab:app-error-ace-by-gcm-full} for baseline errors)}
    \begin{subtable}{0.49\linewidth}
        \centering
        \caption{Missing edges}
        \resizebox{\linewidth}{!}{%
        \begin{tabular}{c|ccc|ccc|ccc}
\toprule
\textbf{DNN} & \multicolumn{3}{c}{ConvNext-B} & \multicolumn{3}{c}{ResNet50} & \multicolumn{3}{c}{Swin-B} \\
\textbf{$N_E$} & 1 & 2 & 4 & 1 & 2 & 4 & 1 & 2 & 4 \\
\midrule
0 & -0.10 & -0.28 & -0.21 & -0.17 & -0.27 & -0.31 & -0.06 & -0.15 & -0.15 \\
1 & -0.27 & 0.03 & 0.55 & -0.14 & 0.06 & 0.38 & -0.12 & 0.24 & 0.82 \\
2 & 0.06 & 0.11 & 0.31 & 0.15 & 0.24 & 0.64 & 0.06 & 0.12 & 0.33 \\
3 & 0.19 & 0.31 & 0.49 & 0.21 & 0.41 & 0.55 & 0.19 & 0.37 & 0.55 \\
4 & 0.08 & 0.09 & 0.06 & 0.09 & 0.06 & -0.04 & 0.09 & 0.07 & 0.00 \\
5 & -0.04 & 0.02 & 0.08 & -0.05 & 0.16 & 0.32 & -0.01 & 0.14 & 0.26 \\
6 & 0.24 & 0.50 & 0.77 & 0.15 & 0.35 & 0.80 & 0.22 & 0.53 & 0.91 \\
7 & 0.12 & 0.16 & 0.23 & -0.08 & -0.12 & -0.02 & 0.02 & 0.08 & 0.01 \\
8 & 0.29 & 0.64 & 1.6 & 0.21 & 0.44 & 0.83 & 0.29 & 0.60 & 1.3 \\
9 & 0.01 & 0.06 & 0.46 & 0.08 & 0.22 & 0.64 & 0.05 & 0.06 & 0.39 \\
\midrule
Mean & 0.06 & 0.17 & 0.44 & 0.04 & 0.16 & 0.38 & 0.07 & 0.21 & 0.44 \\
Std & 0.50 & 0.79 & 1.3 & 0.45 & 0.73 & 1.0 & 0.44 & 0.78 & 1.2 \\
\bottomrule
\end{tabular}
        }%
        \label{tab:app-error-ace-missing-edges-delta-full}
    \end{subtable}
    \hfill
    \begin{subtable}{0.49\linewidth}
        \centering
        \caption{Added edges}
        \resizebox{\linewidth}{!}{%
        \begin{tabular}{c|ccc|ccc|ccc}
\toprule
\textbf{DNN} & \multicolumn{3}{c}{ConvNext-B} & \multicolumn{3}{c}{ResNet50} & \multicolumn{3}{c}{Swin-B} \\
\textbf{$N_E$} & 1 & 2 & 4 & 1 & 2 & 4 & 1 & 2 & 4 \\
\midrule
0 & -0.01 & -0.02 & -0.02 & 0.01 & -0.01 & -0.00 & 0.00 & 0.00 & 0.01 \\
1 & 0.06 & 0.11 & 0.10 & 0.05 & 0.12 & 0.10 & 0.07 & 0.12 & 0.10 \\
2 & 0.01 & 0.00 & -0.07 & 0.04 & 0.06 & -0.03 & 0.02 & 0.03 & -0.03 \\
3 & 0.02 & 0.04 & 0.04 & 0.03 & 0.04 & -0.00 & 0.04 & 0.07 & 0.06 \\
4 & -0.00 & 0.01 & 0.01 & -0.02 & -0.04 & -0.02 & -0.03 & -0.05 & -0.03 \\
5 & -0.03 & -0.02 & -0.04 & -0.02 & -0.01 & -0.07 & -0.07 & -0.05 & -0.02 \\
6 & 0.00 & -0.03 & -0.03 & -0.01 & -0.04 & -0.03 & 0.01 & -0.01 & -0.01 \\
7 & 0.12 & 0.11 & 0.14 & 0.03 & 0.01 & 0.04 & 0.07 & 0.06 & 0.09 \\
8 & 0.06 & 0.07 & 0.17 & 0.04 & 0.04 & 0.17 & 0.01 & 0.01 & 0.11 \\
9 & 0.03 & 0.05 & 0.04 & 0.05 & 0.06 & 0.02 & 0.06 & 0.07 & 0.05 \\
\midrule
Mean & 0.03 & 0.03 & 0.04 & 0.02 & 0.02 & 0.02 & 0.02 & 0.03 & 0.03 \\
Std & 0.11 & 0.13 & 0.13 & 0.10 & 0.13 & 0.14 & 0.11 & 0.13 & 0.11 \\
\bottomrule
\end{tabular}
        }%
        \label{tab:app-error-ace-added-edges-delta-full}
    \end{subtable}
    \label{tab:app-error-ace-edge-errors-full}
\end{figure*}

\section{Causal identification example}
\label{app:example-adjustment}
\begin{figure*}[t!]
    \centering
    \begin{minipage}[ht]{0.55\linewidth}
        \centering
        \resizebox{\linewidth}{!}{%
        \includegraphics[width=\columnwidth]{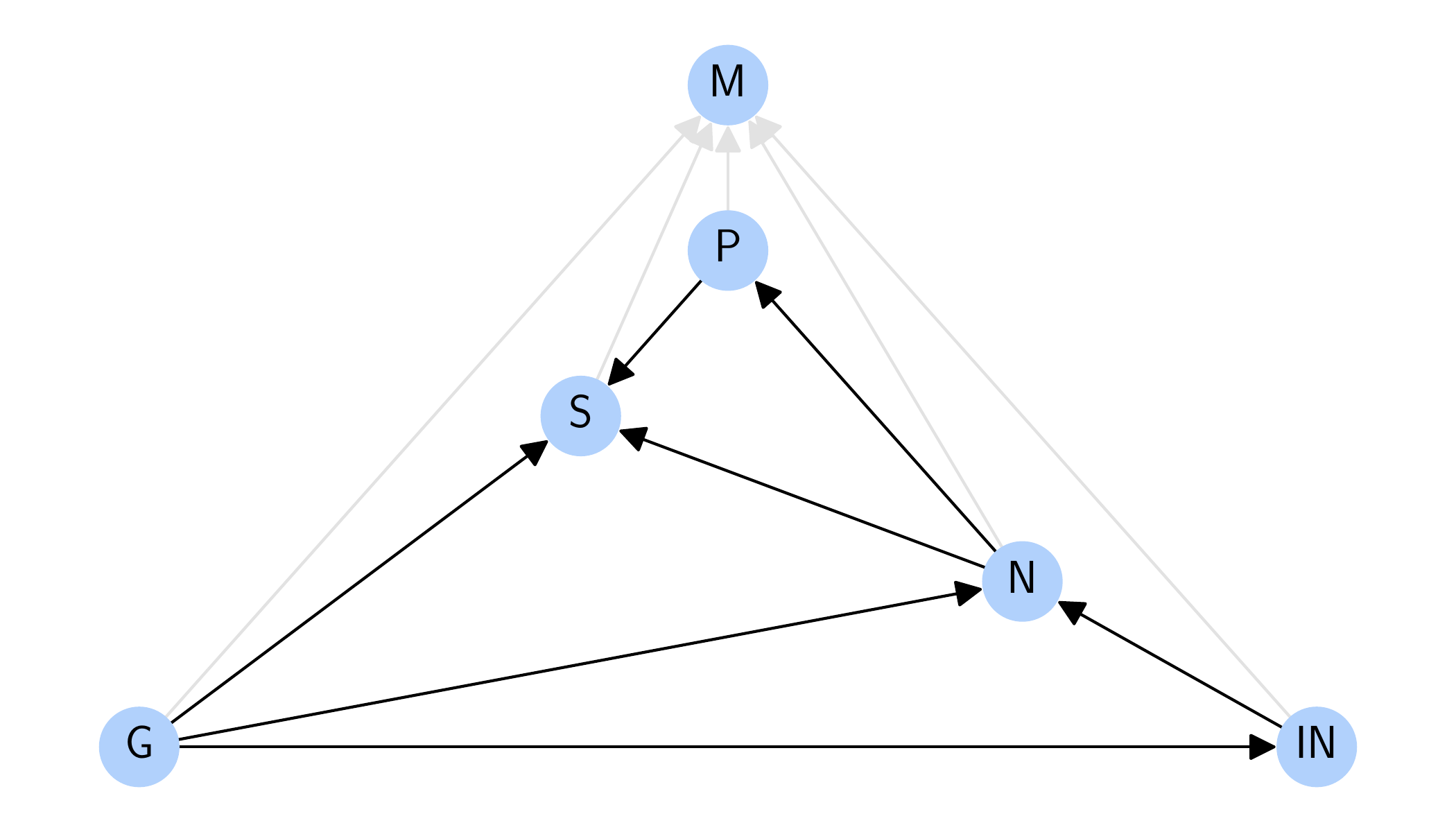}
        }%
        \captionof{figure}{Causal DAG for GCM 0 including the node $M$ corresponding to the correctness metric.}
        \label{fig:gcm-0-backdoor}
    \end{minipage}%
    \hfill
    \begin{minipage}[ht]{0.38\linewidth}
        \centering
        \captionof{table}{Backdoor variables obtained via causal identification for GCM 0 in Figure~\ref{fig:gcm-0-backdoor}}
        \resizebox{0.65\linewidth}{!}{%
        \begin{tabular}{cc}
            \toprule
            Variable & Adjustment Set \\
            \midrule
            G & [] \\
            IN & [G] \\
            N & [G, IN] \\
            P & [N] \\
            S & [G, P, N] \\
            \bottomrule
        \end{tabular}  
        }%
        \label{tab:backdoors}
    \end{minipage}
\end{figure*}

We provide here an example of the identification process for variables in GCM 0 shown in full in Figure~\ref{fig:gcm-0-backdoor}.  Table~\ref{tab:backdoors} provides the set of variables identified as the adjustment set $\gW$ using the backdoor criterion to be included as input to the $\hataceMV$ estimator (\eg $\hat{\mu}(w,v)$ from Sec.~\ref{sub:ace}).

\section{Sample images from GCMs}
\label{app:sample-images}
\begin{figure*}[h]
    \centering
    \begin{subfigure}[b]{\linewidth}
        \includegraphics[width=\linewidth]{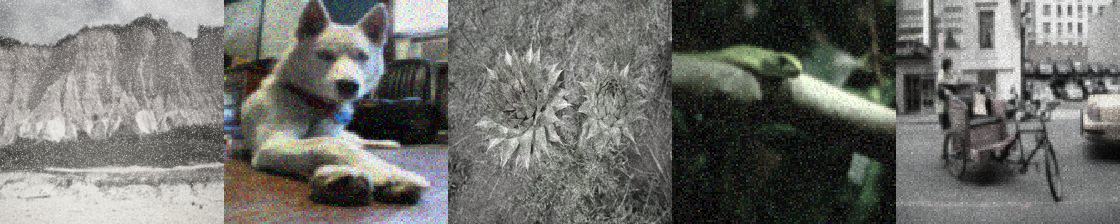} %\\
        \caption{GCM 0}
        \label{fig:app-dag-00-images}
    \end{subfigure}
    \begin{subfigure}[b]{\linewidth}
        \includegraphics[width=\linewidth]{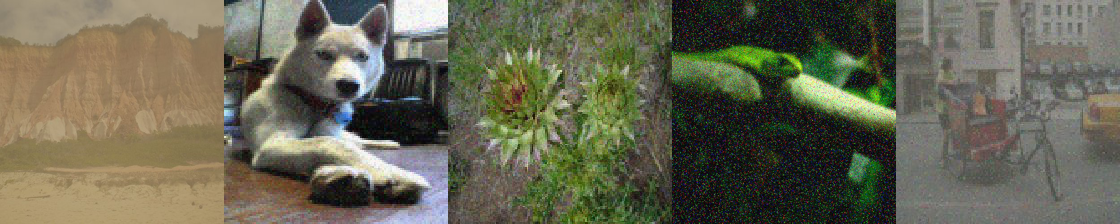} %\\
        \caption{GCM 1}
        \label{fig:app-dag-01-images}
    \end{subfigure}
    \begin{subfigure}[b]{\linewidth}
        \includegraphics[width=\linewidth]{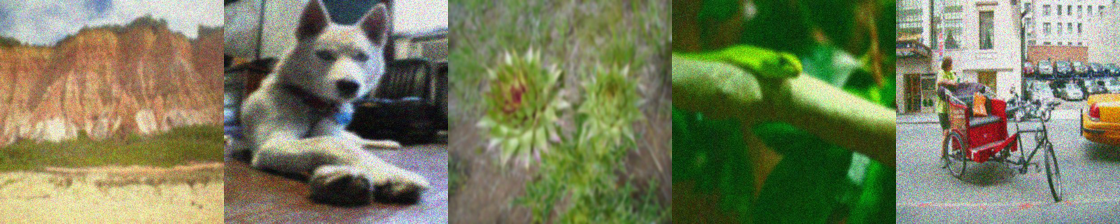} %\\
        \caption{GCM 2}
        \label{fig:app-dag-02-images}
    \end{subfigure}
    \caption{A random subset of images rendered according to the settings sampled from each corresponding GCM.  The nature and severity of imaging conditions are diverse across GCMs illustrating how changes in the GCM DAG topology can have a measurable impact on the image quality. \textit{Images best viewed digitally.}}
    \label{fig:app-dag-images}
\end{figure*}

Figure~\ref{fig:app-dag-images} shows examples of images sampled from the GCMs used for the experiments in Sections~\ref{sub:exp-cdra} and~\ref{sub:exp-dag-err}.  The examples in Figure~\ref{fig:app-dag-images} are sampled randomly from the full set of 50k images and rendered using the process described in Section~\ref{sec:synthetic-data}.  Note that the imaging conditions here are more complex than if only a single factor is applied (\ie the approach used by the common corruptions framework), yet the images are still interpretable.  The low mean accuracies in Table~\ref{tab:app-mean-acc-full} show that these compounded corruptions have a significant impact on DNN performance despite the adequate interpretability of the images.

\section{Rendered corruptions}
\label{app:sampling_corruptions}
The following describes how values from the GCM in Section~\ref{sub:exp-tasks} were sampled, normalized, and used to render CLEVR and MOVi-C variants. Because the imaging factors correspond to continuous values in Blender, we specify functional relationships between various factors in the GCM. For each factor $A$, the normalized factor value is first computed as
\begin{align}
    Z &= \sum_{A^\prime \in pa(A)} \alpha_{(A^\prime,A)} \cdot V_{A^\prime} + U_A \\
    V_A &= \beta_{A}^{-1}(f(Z))
\end{align}
where $\alpha_{(A^\prime, A)}$ is the associated directed edge weight for the edge $(A^\prime, A)$, $V_{A^\prime}$ the sampled value for the parent factor $A^\prime$, $f: \R \rightarrow [0, 1]$ a normalization function, and $\beta_{A}^{-1}(Z)$ the inverse CDF of a random variable $Z \sim \beta(a_A, b_A)$, and $U_A \sim N(0, \sigma_A)$ is an exogenous noise term. For experiments in Sec. 4.4, the edge weights were each randomly sampled from $\gU(-1, 1)$.

\begin{table}[thbp!]
    \centering
    \caption{Directed edge weights $\alpha_{A_i A_j}$ for each edge in the GCM of Section~\ref{sub:exp-tasks}. Edge weights were sampled from $\gU(-1,1)$.}
    \resizebox{0.2\columnwidth}{!}{%
    \begin{tabular}{ll|c}
    \toprule
    $A_i$ & $A_j$ & Weight\\
    \midrule
    L & E & -0.223\\
    L & D & -0.800\\
    E & D & 0.800\\
    E & N & -0.322\\
    D & N & -0.909\\
    \bottomrule
    \end{tabular}
    }%
    \label{tab:hyperparameters_edges}
\end{table}

\begin{table*}[thbp!]
    \centering
    \caption{Hyperparameters for each factor to sample corruption severities in Section~\ref{sub:exp-tasks}}
    \resizebox{0.9\textwidth}{!}{%
    \begin{tabular}{ll|cccccccc}
    \toprule
    Factor & Render Setting & Corruption Type & Min & Max & Nominal & $a_A$ & $b_A$ &  $f(\cdot)$ & $\sigma_A$\\
    \midrule
    L & Light Level & Centered &  0.25 & 1.5 & 1 &  2 & 2 & $(1 + tanh) / 2$ & 1\\
    E & Exposure & Centered & -2 & 2 & 0 & 3 & 3 & $(1 + tanh) / 2$ & 0.1\\
    D & Defocus (via F-stop) & Decreasing & 0.01 & 0.2 & - & 2 & 5 & $(1 + tanh) / 2$ & 0.1\\
    N & Noise (via render cycles) & Decreasing & 10 & 300 & - & 1 & 1 & $(1 + tanh) / 2$ & 0.1\\
    \bottomrule
    \end{tabular}
    \label{tab:hyperparameters_nodes}    
    }%
\end{table*}

Normalized factor values are mapped back to Blender settings to enable physics-based scene rendering. The final Blender setting is calculated for each factor. For \textit{increasing} factors, the normalized $V_A$ is rescaled to $[\min_A, \max_A]$ where $\min_A,~\max_A$ correspond to the minimum, maximum settings to be allowed in Blender respectively. For \textit{decreasing} factors, the value $1 - V_A$ is computed first and then mapped to $[\min_A, \max_A]$. For \textit{centered} corruptions, the value is first rescaled to $V'_A \in [\min_A, \max_A]$ and then adjusted according to $\frac{\left|V'_A - n_A\right|}{\max\left(\left|\min_A - n_A\right|, \left|\max_A - n_A\right|\right)}$ where $n_A$ is the nominal value that yields no/minimal effect of that attribute on the rendered image. Normalization settings are given in Tables~\ref{tab:hyperparameters_edges} and~\ref{tab:hyperparameters_nodes}. 

\section{CDRA for optical flow}
\label{app:optical-flow}
For evaluating the robustness of optical flow methods, we also \textit{render} a new variant of the Kubric MOVi-C benchmark dataset~\cite{greff2021kubric} using the same GCM DAG in Figure~\ref{fig:ocl-flow-scm} and according to the process described in Appendix~\ref{app:sampling_corruptions}.  We evaluate three top performing baselines FlowNetC~\cite{dosovitskiy2015flownet}, PWCNet~\cite{sun2018pwc}, and RAFT~\cite{teed2020raft} and use the standard average Endpoint Error ($EPE$) for measuring overall performance and use CDRA to estimate $ACE_{EPE}$.

\begin{table}[thbp!]
    \centering
    \caption{Comparison of $ACE_{EPE}$ for multiple optical flow baseline algorithms.}
    \resizebox{0.5\columnwidth}{!}{%
    \begin{tabular}{lcccc|c}
        \toprule
         & \multicolumn{4}{c}{$ACE_{EPE}(\cdot)$} & \\
        DNN / Factor & C & E & F & L & EPE \\
        \midrule
        FlowNetC & \bfseries 18.7 & \bfseries -19.4 & \bfseries 57.7 & \bfseries -1.19 & 10.7\\
        PWCNet & 18.8 & -19.5 & 58.0 & -1.17 & 10.3 \\
        RAFT & 18.8 &  -19.6 & 58.0 & -1.18 & \bfseries 10.3\\
        \bottomrule
    \end{tabular}
    }%
    \label{tab:of-ace-epe}    
\end{table}

\noindent\textbf{Results:~}
The results in Table~\ref{tab:of-ace-epe} show little difference in $ACE_{EPE}$ and $EPE$ between algorithms but large differences in $ACE_{EPE}$ for each factor. Whereas average $EPE$ is similar across all models, we observe that noise and defocus have much larger adverse effects on $EPE$ relative to the other factors. 

\noindent\textbf{Discussion:~}
These results indicate that CDRA exposes sensitivities of the optical flow models not observed when measuring only average performance on the dataset. These insights enable more targeted downstream robustness interventions in the DNN architecture design, training data collection, or optimization strategy to address these sensitivities.

\section{Compute resources}
\label{app:compute}
All experiments can be executed using a single, locally-hosted NVIDIA A40 GPU with 48GB of memory. Data generation with both the compositing and rendering methods required a single A40 GPU as well. The causal effect estimation and root cause analysis were conducted using a single laptop CPU.

\end{document}